\documentclass[10pt,twocolumn,letterpaper]{article}

\usepackage{cvpr}              % To produce the CAMERA-READY version
\usepackage{tabularx}
\usepackage{multirow}
\usepackage{xcolor}
\usepackage{algorithm}
\usepackage{algpseudocode}
\usepackage{graphicx}
\definecolor{cvprblue}{rgb}{0.21,0.49,0.74}
\usepackage[pagebackref,breaklinks,colorlinks,allcolors=cvprblue]{hyperref}

\newcommand{\best}[1]{\textbf{\textcolor{red}{#1}}}
\newcommand{\second}[1]{\underline{\textcolor{blue}{#1}}}
\newcommand{\third}[1]{\textit{\textcolor{green!60!black}{#1}}}

\def\paperID{4415} % *** Enter the Paper ID here
\def\confName{CVPR}
\def\confYear{2025}

\title{SlimDiffSR: Toward Lightweight and Efficient Remote Sensing Image Super-Resolution via Diffusion Model Distillation}

\author{Ce Wang, Zhenyu Hu, Wanjie Sun\thanks{Corresponding author}\\
School of Remote Sensing and Information Engineering, Wuhan University\\ Wuhan 430079, China\\
\tt\small {\{cewang, zhenyuhu, sunwanjie\}@whu.edu.cn}
}

\begin{document}
\maketitle
\begin{abstract}
Diffusion models have recently achieved remarkable performance in image super-resolution (SR), but their high computational cost limits practical deployment in remote sensing applications. To address this issue, we propose SlimDiffSR, a lightweight and efficient diffusion-based framework for real-world remote sensing image super-resolution. Unlike existing single-step diffusion methods that rely on fixed timesteps, we first introduce an uncertainty-guided timestep assignment strategy to construct a stronger single-step teacher model, where reconstruction difficulty is explicitly linked to diffusion timesteps, enabling adaptive generative strength. Building upon this teacher, we further present a structured pruning strategy tailored to remote sensing imagery, which systematically removes redundant semantic modules and replaces standard operations with lightweight designs, including frequency-separable convolution, direction-separable convolution, and a query-driven global aggregation module. These components explicitly exploit the unique characteristics of remote sensing data, such as sparse high-frequency details, strong directional patterns, and long-range spatial dependencies. To enhance knowledge transfer, we incorporate Maximum Mean Discrepancy (MMD) into the distillation process to align feature distributions between the teacher and student models. Extensive experiments on multiple remote sensing benchmarks demonstrate that SlimDiffSR achieves a favorable balance between efficiency and reconstruction quality. In particular, it attains up to $200\times$ inference acceleration and a $20\times$ reduction in model parameters compared with multi-step diffusion models, while achieving competitive perceptual quality and clearly outperforming existing lightweight diffusion baselines in efficiency. The code is available at: \url{https://github.com/wwangcece/SlimDiffSR}.
\end{abstract}
\section{Introduction}
\label{sec:introduction}

High-resolution (HR) remote sensing imagery is critical for a wide range of Earth observation and monitoring applications, including but not limited to weather forecasting \cite{venter2020hyperlocal}, object detection \cite{li2020object}, and land use and land cover mapping \cite{tong2020land}. However, the spatial resolution of many satellite images is inherently constrained by hardware limitations and transmission bandwidth, often resulting in insufficient detail for reliable interpretation and analysis \cite{huang2013spatial}. To address this limitation, super-resolution (SR) techniques aim to enhance the spatial fidelity of low-resolution (LR) observations through data-driven reconstruction, enabling the generation of HR imagery in a cost-effective manner.

Over the past decades, a variety of SR approaches have been developed \cite{wang2020deep}, including reconstruction-based methods, sparse representation-based methods, and regression-based methods. More recently, deep learning has significantly advanced SR performance by leveraging large-scale datasets to learn complex LR-to-HR mapping. Early deep SR research primarily focused on network architecture design \cite{zhang2018residual,kim2016accurate,chen2023activating}, loss functions \cite{johnson2016perceptual,ledig2017photo}, and optimization strategies \cite{wang2020deep}, achieving substantial improvements in distortion-oriented metrics such as peak signal-to-noise ratio (PSNR). Subsequently, generative adversarial networks (GANs) \cite{goodfellow2020generative,zhang2023real} were introduced to enhance perceptual quality by encouraging the generation of realistic textures through adversarial training. Despite their success, GAN-based methods often suffer from training instability, mode collapse, and the introduction of undesirable visual artifacts \cite{liang2022details}. In recent years, diffusion models (DMs) \cite{ho2020denoising} have demonstrated remarkable success in high-fidelity image generation and have been extended to a variety of low-level vision tasks, including image-to-image translation and image restoration. Their application to SR has shown promising results, offering improved perceptual quality, enhanced training stability, and more diverse reconstruction outcomes \cite{yue2023resshift,wang2024exploiting,lin2024diffbir}.

Despite these advantages, diffusion-based SR methods remain computationally expensive due to their iterative sampling procedures and large model sizes. For instance, a typical multi-step diffusion model \cite{wang2024exploiting} may require more than 100 seconds to process a single $1024 \times 1024$ image on an RTX 3090 GPU. Such computational cost poses a significant barrier to large-scale deployment in remote sensing applications, where massive volumes of imagery must be processed efficiently. To mitigate this issue, recent studies have focused on developing lightweight diffusion frameworks, which can be broadly categorized into three directions. First, task-specific diffusion formulations have been proposed for SR and image restoration \cite{yue2023resshift, liu2024residual}, aiming to exploit degraded inputs more effectively while reducing the number of denoising steps. Second, advanced ordinary differential equation (ODE) and stochastic differential equation (SDE) solvers have been introduced to accelerate the sampling process \cite{song2020denoising, lu2022dpm}. Third, model compression techniques, such as knowledge distillation and structured pruning, have been employed to simultaneously reduce sampling steps and model complexity \cite{chen2025adversarial, kim2024bk}.

Despite these advances, several challenges remain when applying lightweight diffusion models to remote sensing SR:

\begin{enumerate}
	\item \textbf{Fixed timestep}: Existing single-step diffusion models typically adopt a fixed timestep \cite{wu2024one, chen2025adversarial}, ignoring the fact that different timesteps in multi-step diffusion pretraining provide distinct guidance signals for the denoising process.
	
	\item \textbf{Suboptimal pruning strategies}: Current pruning methods are largely adapted from natural image generation tasks \cite{kim2024bk}, without accounting for the unique characteristics of SR and remote sensing imagery.
	
	\item \textbf{Limited distillation objectives}: Most existing distillation approaches rely on mean squared error (MSE) to align intermediate features between teacher and student networks \cite{kim2024bk}, which only captures first-order statistics and fails to model distribution-level discrepancies.
\end{enumerate}

\begin{figure}[htb]
	\centering
	\includegraphics[width=\linewidth]{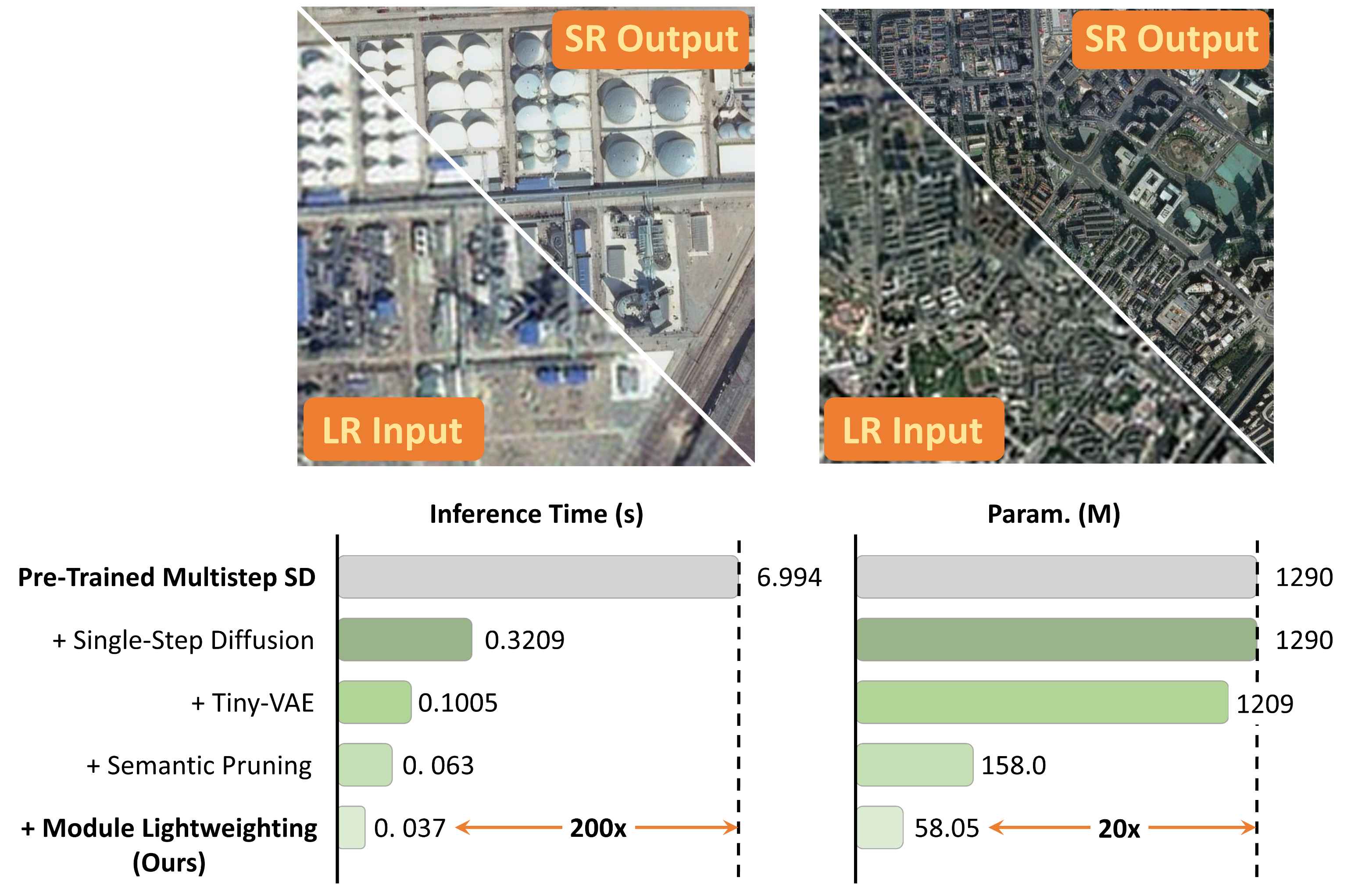}
	\caption{Our method achieves a substantial reduction in both inference time and model size. By constructing a single-step diffusion model and further incorporating structured pruning techniques, the final lightweight model attains up to 200$\times$ inference acceleration and a 20$\times$ reduction in the number of parameters, while maintaining competitive restoration performance.}
	\label{fig:component_efficiency}
\end{figure}

To address these challenges, we propose SlimDiffSR, a lightweight and efficient diffusion-based framework for real-world remote sensing image super-resolution. As illustrated in Fig. \ref{fig:component_efficiency}, the proposed method achieves significant improvements in both inference efficiency and model compactness. The overall framework consists of two stages: teacher model training and student model distillation.

In the first stage, we construct an unpruned single-step diffusion-based SR network as the teacher model. Although existing single-step diffusion-based methods, such as OSEDiff \cite{wu2024one} and S3Diff \cite{zhang2024degradation}, demonstrate promising runtime efficiency, they rely on fixed diffusion timestep and thus fail to fully exploit timestep-dependent guidance learned during diffusion pretraining. Our prior studies \cite{wang2025timestep} have shown that different inputs may require different timesteps to achieve better restoration quality due to varying degradation levels. Motivated by this observation, we propose an uncertainty-guided single-step diffusion model for remote sensing image SR as the teacher model. Specifically, we establish a correspondence between reconstruction uncertainty and diffusion timesteps. An uncertainty estimator is first trained in the latent space using an uncertainty-aware objective \cite{ning2021uncertainty} to predict reconstruction difficulty. A timestep inversion strategy is then introduced to map uncertainty scores to appropriate diffusion timesteps, enabling dynamic timestep assignment for different LR inputs.

In the second stage, the teacher model is pruned and distilled to improve inference efficiency while preserving reconstruction performance. This stage consists of two steps: structural pruning and distillation training. For the pruning strategy, the denoising network is decomposed into removable and replaceable modules. Since LR inputs already contain substantial semantic information, and deeper UNet layers of the diffusion model primarily capture high-level semantics \cite{voynov2023p+}, these components are considered redundant for SR and can be removed. Second, considering the characteristics of remote sensing imagery, we redesign the remaining modules to better match the data properties. Remote sensing images typically contain large homogeneous regions with limited low-frequency variation, while critical information is concentrated in high-frequency structures such as edges and object boundaries. To exploit this, we introduce frequency-separable convolution to emphasize informative high-frequency components while reducing redundant computation. In addition, remote sensing scenes often exhibit strong directional patterns (e.g., roads and rivers). To capture such anisotropic structures efficiently, we design direction-separable convolution. Furthermore, due to the large spatial extent of remote sensing images, long-range dependencies are important but costly to model with standard self-attention. Therefore, we propose a query-driven global aggregation module to approximate global context in a more efficient manner. These designs replace standard convolution and self-attention layers, improving efficiency while preserving task-relevant information. For distillation, a multi-scale feature supervision strategy is adopted to transfer knowledge from the teacher to the student network. To address the distribution mismatch between their features, we incorporate the Maximum Mean Discrepancy (MMD) loss \cite{gretton2012kernel} alongside the MSE loss to enforce both point-wise and distribution-level alignment.

\begin{figure}[htb]
	\centering
	\includegraphics[width=\linewidth]{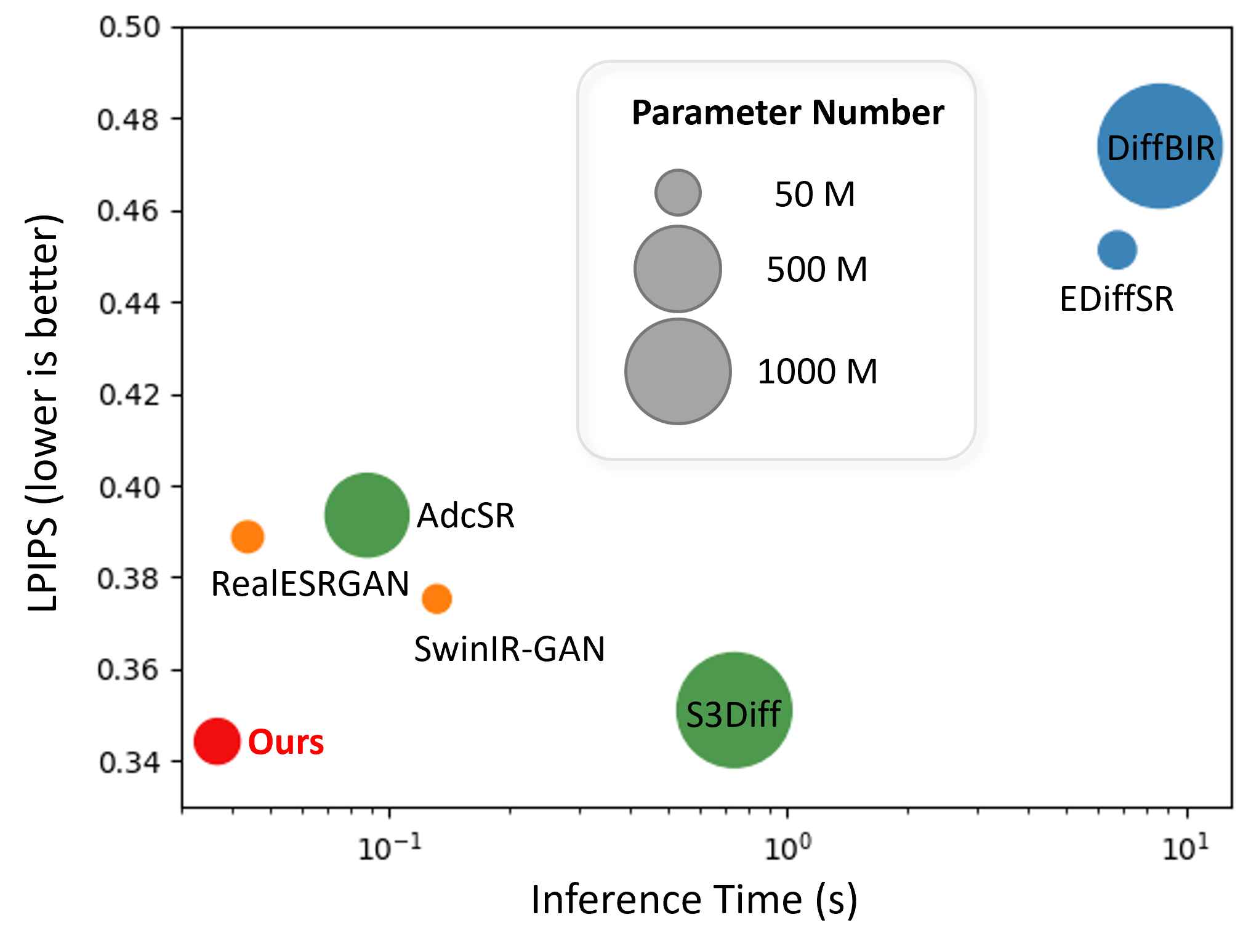}
	\caption{Performance and efficiency comparison among different remote sensing image SR methods for processing a single 512 $\times$ 512 image on an RTX 3090 GPU. Red bubbles denote multi-step diffusion-model-based methods, green bubbles denote single-step diffusion-model-based methods, and yellow bubbles denote GAN-based methods. Our method achieves the fastest inference speed and the best LPIPS score while maintaining a relatively small model size.}
	\label{fig:compare_efficiency}
\end{figure}

Unlike existing lightweight diffusion-based SR methods that directly approximate multi-step inference with fixed designs, this work addresses a key limitation in single-step diffusion for remote sensing SR, i.e., the lack of adaptive generative strength under varying degradation levels. To this end, we introduce an uncertainty-guided timestep assignment strategy to construct a stronger single-step teacher model, where reconstruction difficulty is explicitly linked to diffusion timesteps. Building upon this teacher, we further develop a lightweight student model through RS-aware structural redesign. Instead of applying generic pruning strategies, we exploit the characteristics of remote sensing imagery to remove redundant semantic modules and replace remaining components with efficient alternatives. This teacher–student design enables us to improve reconstruction quality through adaptive diffusion guidance, while achieving high inference efficiency via lightweight model construction. As shown in Fig. \ref{fig:compare_efficiency}, while maintaining a compact model size and strong reconstruction quality, our lightweight model not only significantly outperforms existing multi-step and single-step DMs-based methods in inference speed, but also surpasses GAN-based approaches.

To summarize, the main contributions of this work are as follows:
\begin{enumerate}
	\item We identify the limitation of fixed-timestep selection in existing single-step diffusion SR models and propose an uncertainty-guided timestep assignment strategy. By explicitly linking reconstruction difficulty with diffusion timesteps, the proposed method enables adaptive generative strength and improves the effectiveness of single-step diffusion for remote sensing SR.
	
	\item We show that effective model compression for remote sensing SR requires RS-specific structural redesign rather than generic pruning. Based on this insight, we remove redundant deep semantic modules and introduce lightweight operators, including frequency-separable convolution, direction-separable convolution, and query-driven global aggregation, which better match the characteristics of remote sensing imagery.
	
	\item We develop a two-stage teacher--student diffusion framework that decouples quality enhancement from efficient deployment. The adaptive teacher provides strong reconstruction supervision, while the lightweight student inherits this capability through distillation and performs fixed single-step inference without additional timestep prediction overhead.
	
	\item We incorporate MMD into the distillation process to complement conventional MSE supervision, enabling alignment of both point-wise features and their underlying distributions, which leads to improved perceptual quality and reconstruction fidelity.
\end{enumerate}

\section{Related Work}
\subsection{Remote Sensing Image Super-Resolution}

Early research on remote sensing image super-resolution (SR) predominantly relied on convolutional neural networks (CNNs). Representative models such as SRCNN \cite{dong2014learning} and VDSR \cite{kim2016accurate} were directly applied to satellite imagery, demonstrating the effectiveness of deep learning for enhancing both panchromatic and multispectral data. For example, Tuna \textit{et al.} \cite{tuna2018single} showed that VDSR performs well on high-resolution images from SPOT and Pleiades sensors. However, these early models are relatively shallow and lack sufficient contextual modeling capability, limiting their ability to capture complex spatial patterns in remote sensing scenes.

To address this limitation, deeper and task-specific CNN-based models have been proposed. The RS-DRL network \cite{granger2017comparing} introduces residual learning for Sentinel-2 data, while the Deep Memory Connected Network (DMCN) \cite{xu2018high} incorporates local and global memory mechanisms to enhance multi-level feature representation. Although these methods achieve strong performance in distortion-oriented metrics such as PSNR and SSIM, they often produce over-smoothed textures and fail to preserve fine details.

To improve perceptual quality, GANs have been introduced into remote sensing SR. Models such as ESRGAN \cite{lanaras2018super}, EEGAN \cite{chen2022super}, and DRGAN \cite{ma2019super} employ adversarial training and perceptual losses to generate sharper and more realistic images. In addition, task-specific variants such as SRAGAN \cite{li2021single} are designed to better capture structural characteristics in remote sensing imagery. However, GAN-based methods often suffer from inferior quantitative performance and may introduce artifacts, especially under large scaling factors or cross-sensor scenarios. To mitigate these issues, subsequent studies have explored hybrid loss functions \cite{liu2020super} and improved discriminator designs \cite{zhang2022single} to balance perceptual quality and reconstruction fidelity.

\subsection{SR Diffusion Models}

Diffusion models (DMs) have recently emerged as a powerful paradigm for image SR, initially in the domain of natural images. Early approaches such as SR3 \cite{saharia2022image} and SRDiff \cite{li2022srdiff} train diffusion models from scratch to reconstruct HR images from LR inputs, but require substantial computational resources and large-scale paired datasets. To alleviate this limitation, recent methods leverage pretrained text-to-image diffusion models as strong generative priors. For example, DiffBIR \cite{lin2024diffbir} and StableSR \cite{wang2024exploiting} utilize pretrained Stable Diffusion (SD) \cite{rombach2022high} to guide texture synthesis, demonstrating that pretrained generative knowledge can be effectively transferred to SR tasks.

Despite their strong performance, pretrained diffusion models are characterized by large parameter sizes and slow inference speeds, which hinder practical deployment. To improve efficiency, recent studies have explored lightweight diffusion techniques from three main perspectives. First, task-specific diffusion formulations, such as ResShift \cite{yue2023resshift} and RDDM \cite{liu2024residual}, aim to reduce the number of denoising steps. Second, efficient ODE/SDE-based samplers \cite{song2020denoising, lu2022dpm} are developed to accelerate the sampling process. Third, model compression techniques, including knowledge distillation and structured pruning, are employed to jointly reduce sampling steps and model complexity. For instance, OSEDiff \cite{wu2024one} and S3Diff \cite{zhang2024degradation} achieve single-step inference via distillation, while AdcSR \cite{chen2025adversarial} further reduces model size through structured pruning.

In the domain of remote sensing, SR poses unique challenges due to complex textures, large scale variations, and sensor-specific imaging characteristics. Early DMs-based SR methods for remote sensing typically relied on training models from scratch using domain-specific datasets. For example, EDiffSR \cite{xiao2023ediffsr} introduces an efficient diffusion probabilistic framework tailored for remote sensing SR, incorporating an Efficient Activation Network for noise prediction and a Conditional Prior Enhancement Module to strengthen conditioning cues. More recent studies have attempted to bridge the gap between natural-image diffusion foundation models and remote sensing applications. SGDM \cite{wang2025semantic}, for instance, leverages pretrained SD models to perform $16\times$ and $32\times$ SR for remote sensing imagery by disentangling style and content representations and incorporating vector map priors to enhance semantic consistency. Likewise, DiffusionSat \cite{khanna2023diffusionsat}, the first diffusion foundation model for remote sensing, combines pretrained SD with auxiliary metadata (e.g., geographic coordinates) to support temporal SR and image inpainting, demonstrating superior performance over GAN-based approaches.

Despite these advances, existing diffusion-based SR methods for remote sensing remain computationally expensive and are not well suited for large-scale deployment. Meanwhile, lightweight diffusion techniques developed for natural images do not fully consider the unique characteristics of remote sensing data, as discussed in Section~\ref{sec:introduction}. Therefore, there remains a need for a dedicated lightweight diffusion framework tailored to remote sensing image SR.

\section{Method}
\subsection{Overview}
\begin{figure}[htb]
	\centering
	\includegraphics[width=\linewidth]{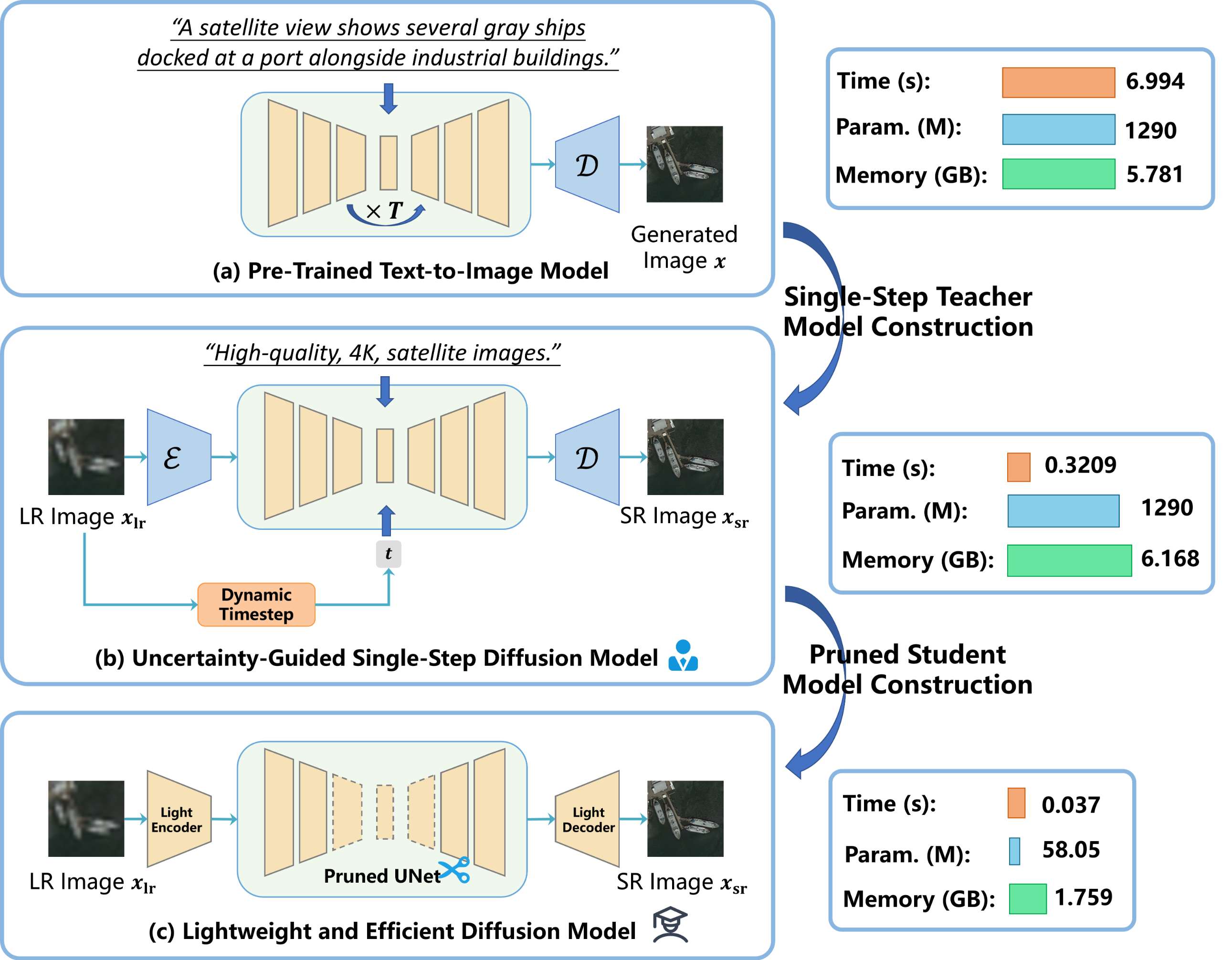}
	\caption{Overview of the proposed SlimDiffSR framework. In the first stage, an uncertainty-guided single-step diffusion teacher is constructed to provide high-quality supervision. In the second stage, the adaptive behavior of the teacher is distilled into a lightweight student model, where redundant modules are removed or replaced for efficient fixed single-step inference.}
	\label{fig:progr_pruned}
\end{figure}

The overall framework of SlimDiffSR is illustrated in Fig. \ref{fig:progr_pruned}. The goal of this work is to develop a lightweight diffusion-based model for remote sensing image super-resolution that achieves high reconstruction quality while maintaining efficient inference. To this end, we adopt a two-stage teacher–student framework. In the first stage, we construct a strong single-step diffusion teacher model to provide high-quality supervision. Unlike existing single-step diffusion methods that rely on fixed timesteps, we introduce an uncertainty-guided timestep assignment strategy to enhance the reconstruction capability of the teacher model under varying degradation levels. In the second stage, we focus on efficiency by designing a lightweight student model through RS-aware structural pruning and module replacement tailored to remote sensing imagery.

Importantly, the adaptive timestep mechanism is only used during teacher model construction. The final student model operates with a fixed single-step inference, introducing no additional computational overhead. The key idea is that the adaptive reconstruction behavior introduced by timestep variation in the teacher model can be transferred to the student via distillation. As a result, the student implicitly learns to approximate the effect of adaptive timestep selection, without explicitly requiring timestep prediction during inference. Through this design, the proposed framework effectively decouples reconstruction quality and inference efficiency, enabling SlimDiffSR to achieve a favorable balance between performance and efficiency for large-scale remote sensing applications.

\subsection{Single-Step Teacher Model Construction}
The objective of the teacher model is to provide strong reconstruction supervision for the subsequent lightweight student. A key limitation of existing single-step diffusion methods lies in their use of a fixed timestep, which implicitly assumes a uniform degradation level across all inputs. However, in remote sensing SR, degradation varies across scenes and spatial position, making such a fixed design suboptimal.

\begin{figure}[htb]
	\centering
	\includegraphics[width=\linewidth]{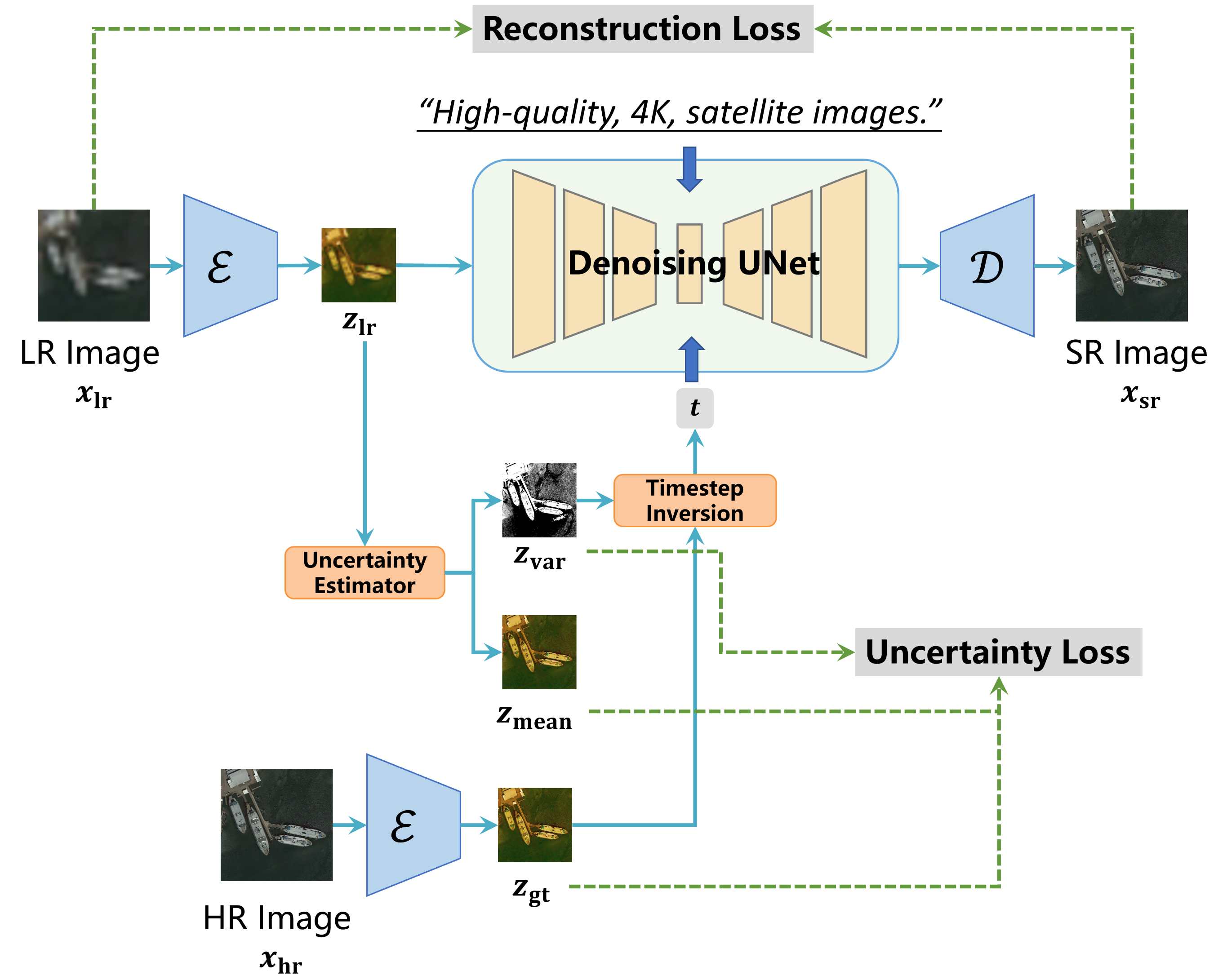}
	\caption{Overview of the uncertainty-guided single-step diffusion model in the first stage.}
	\label{fig:stage1}
\end{figure}

To address this issue, we introduce an uncertainty-guided timestep assignment strategy to construct a stronger single-step teacher model. Instead of treating the timestep as a fixed hyperparameter, we learn an adaptive diffusion guidance level conditioned on reconstruction uncertainty and latent structural information. The uncertainty map provides an important cue for estimating restoration difficulty, while the final timestep is optimized by the teacher restoration objective. As illustrated in Fig.~\ref{fig:stage1}, given an LR input $x_{\mathrm{lr}}$, an uncertainty estimator first predicts an element-wise uncertainty map $z_{\mathrm{var}}$, which reflects reconstruction ambiguity. Then, a timestep inversion strategy estimates the corresponding diffusion timestep $t^*$ by jointly considering uncertainty-aware cues and latent structural information. The predicted timestep is used to guide the UNet denoising process, allowing the teacher model to adapt its generative strength to different inputs and provide stronger supervision than a plain fixed-timestep single-step diffusion SR model.

\subsubsection{Uncertainty Estimation}
According to existing studies on uncertainty learning \cite{ning2021uncertainty}, we model the ground-truth latent $z_{\mathrm{gt}}$ as a Gaussian distribution conditioned on the LR latent $z_{\mathrm{lr}}$:
\begin{equation}
	z_{\mathrm{gt}} = z_{\mathrm{mean}} + \epsilon \cdot \sqrt{z_{\mathrm{var}}}, \quad \epsilon \sim \mathcal{N}(0, \mathrm{I})
	\label{eq:observation_model}
\end{equation}
where $z_{\mathrm{mean}}$ and $z_{\mathrm{var}}$ denote the predicted mean and variance, respectively. Both are estimated by an uncertainty estimator $(z_{\mathrm{mean}}, z_{\mathrm{var}}) = \mathrm{UE}_{\phi}(z_{\mathrm{lr}})$. Intuitively, the role of the uncertainty estimator $\mathrm{UE}_{\phi}$ can be decomposed into two complementary components. First, it performs a preliminary restoration of the LR input $z_{\mathrm{lr}}$, producing a mean estimate $z_{\mathrm{mean}}$ that represents the expected HR latent under the predicted distribution in which most degradations are suppressed. Second, it estimates the reconstruction uncertainty for each pixel, yielding a variance map $z_{\mathrm{var}}$. As observed in prior studies \cite{kong2021classsr}, texture-rich regions are generally more difficult to accurately reconstruct, leading to higher uncertainty values for the reconstruction process.

To train the estimator, we maximize the posterior probability of the predicted variance conditioned on the observed LR-HR latent pair. According to Bayes' rule, the posterior can be written as:
\begin{equation}
	p(z_{\mathrm{var}} \mid z_{\mathrm{gt}}, z_{\mathrm{lr}})
	\propto 
	p(z_{\mathrm{gt}} \mid z_{\mathrm{mean}}, z_{\mathrm{var}})
	p(z_{\mathrm{var}})
	\label{eq:posterior_prior}
\end{equation}
where $z_{\mathrm{mean}}$ is implicitly conditioned on $z_{\mathrm{lr}}$. Assuming a Gaussian observation model, the likelihood is given by:
\begin{equation}
	p(z_{\mathrm{gt}} \mid z_{\mathrm{mean}}, z_{\mathrm{var}})
	=
	\frac{1}{\sqrt{2\pi z_{\mathrm{var}}}}
	\exp\left(
	-\frac{\|z_{\mathrm{gt}} - z_{\mathrm{mean}}\|_2^2}{2z_{\mathrm{var}}}
	\right)
	\label{eq:gaussian_likelihood}
\end{equation}

Meanwhile, according to Jeffreys' prior \cite{figueiredo2001adaptive}, the uncertainty prior can be expressed as $p(z_{\mathrm{var}}) \propto \frac{1}{z_{\mathrm{var}}}$. By taking the logarithm of the posterior and removing constant terms, we obtain:
\begin{equation}
	\ln p(z_{\mathrm{var}} \mid z_{\mathrm{gt}}, z_{\mathrm{lr}})
	\propto
	-\frac{\|z_{\mathrm{gt}} - z_{\mathrm{mean}}\|_2^2}{2z_{\mathrm{var}}}
	-\frac{3}{2}\ln z_{\mathrm{var}}
\end{equation}

Minimizing the negative log-posterior leads to the following uncertainty-aware loss. Let $s = \ln z_{\mathrm{var}}$ for numerical stability, then the loss is defined as:
\begin{equation}
	\mathcal{L}_{\text{uncertainty}}
	=
	\frac{1}{2N}
	\sum_{i=1}^{N}
	\left(
	\exp(-s^i)
	\|z_{\mathrm{gt}}^{i} - z_{\mathrm{mean}}^{i}\|_2^2
	+
	3s^i
	\right)
\end{equation}
here, the variance is predicted element-wise over the latent representation, and each latent element is modeled as an independent scalar Gaussian variable. This formulation enables the estimator to adaptively assign higher uncertainty to regions that are more difficult to reconstruct, which subsequently guides the timestep assignment in the diffusion model.

\subsubsection{Timestep Inversion}
Our previous studies have shown that, for single-step diffusion models, inputs with different reconstruction difficulties should be guided by different denoising timesteps \cite{wang2025timestep}. Given the uncertainty map estimated in the previous section, our goal is to determine an appropriate timestep for the teacher model, so that the generative strength of the diffusion model can be adaptively adjusted according to the reconstruction ambiguity.

A straightforward solution is to directly predict the timestep $t$ with a neural network. However, this design is not suitable for end-to-end optimization, because the predicted timestep is used as a discrete index to retrieve the corresponding noise schedule value $\bar{\alpha}_t$. Such an indexing operation is non-differentiable, making it difficult to train the timestep prediction module jointly with the teacher model. Therefore, instead of directly predicting the discrete timestep $t$, we predict a continuous noise-level variable $\bar{\alpha}_{\text{target}}$, which can be regarded as a continuous representation of the diffusion timestep.

To provide a principled target for predicting $\bar{\alpha}_{\text{target}}$, we establish the relationship between reconstruction uncertainty and diffusion noise level through the signal-to-noise ratio (SNR). In the forward diffusion process, the noisy latent at timestep $t$ is formulated as:
\begin{equation}
	z_t = \sqrt{\bar{\alpha}_t} z_0 + \sqrt{1-\bar{\alpha}_t}\epsilon,
	\quad \epsilon \sim \mathcal{N}(0,\mathrm{I})
	\label{eq:diffusion_forward}
\end{equation}
where $\bar{\alpha}_t$ controls the relative proportion of signal and noise. The corresponding SNR is given by:
\begin{equation}
	\mathrm{SNR}_{\mathrm{diff}}(t)
	=
	\frac{\bar{\alpha}_t}{1-\bar{\alpha}_t}
	\label{eq:snr_time}
\end{equation}

Similarly, the uncertainty-based observation model describes reconstruction difficulty through the predicted variance $z_{\mathrm{var}}$. A larger variance indicates higher uncertainty and thus a lower signal-to-noise ratio. Therefore, the uncertainty-based SNR can be written as:
\begin{equation}
	\mathrm{SNR}_{\mathrm{uncertainty}}
	=
	\frac{\sigma_{\mathrm{gt}}^2}{z_{\mathrm{var}}}
	\label{eq:snr_signal}
\end{equation}
where $\sigma_{\mathrm{gt}}^2$ denotes the signal variance of the HR latent distribution. By aligning the SNR of the diffusion process with the SNR implied by the uncertainty model, we obtain:
\begin{equation}
	\bar{\alpha}_{\text{target}}
	=
	\frac{\sigma_{\mathrm{gt}}^2}
	{\sigma_{\mathrm{gt}}^2+z_{\mathrm{var}}}
	\label{eq:alpha_hat_solution}
\end{equation}

The above derivation provides a theoretical tendency that higher reconstruction uncertainty generally corresponds to a lower SNR and thus a larger timestep, which introduces stronger generative guidance. However, this relationship is used as a guiding principle rather than a hard constraint. In practice, the optimal timestep for restoration is also affected by image structure, semantic context embodied in $\sigma_{\mathrm{gt}}^2$. Directly computing Eq.~\ref{eq:alpha_hat_solution} is impractical because $\sigma_{\mathrm{gt}}^2$ is unknown and the Gaussian assumption may not fully describe the complex distribution of real remote sensing imagery. Therefore, we use Eq.~\ref{eq:alpha_hat_solution} as a theoretical guideline and introduce a prediction network $\mathrm{Pred}_{\theta}$ to estimate $\bar{\alpha}_{\text{target}}$ in a data-driven manner:
\begin{equation}
	\bar{\alpha}_{\text{target}}
	=
	\mathrm{Pred}_{\theta}
	\left(
	z_{\mathrm{gt}}, z_{\mathrm{var}}
	\right)
	\label{eq:pred_alpha}
\end{equation}
here, $z_{\mathrm{var}}$ provides uncertainty information, while $z_{\mathrm{gt}}$ is available during teacher model construction and provides latent structural reference information. The predictor is optimized jointly with the teacher model through the final reconstruction and adversarial objectives, allowing it to learn the noise level that yields the best restoration performance. It is worth noting that this prediction module is only used for constructing the teacher model and is not involved in the inference of the final lightweight student model.

After obtaining $\bar{\alpha}_{\text{target}}$, we need to convert it into the corresponding timestep $t^*$. Since $\bar{\alpha}_t$ is determined by the predefined diffusion noise schedule, this conversion can be written as $t^* = \bar{\alpha}^{-1} \left( \bar{\alpha}_{\text{target}} \right)$. To avoid the potential artifacts from neural network-based inversion, we adopt a piecewise linear interpolation scheme based on the precomputed noise schedule. Specifically, we maintain two lookup tables: an ascending sequence of noise levels $\{\bar{\alpha}_i\}_{i=0}^{1000}$ sampled from the original diffusion scheduler and the corresponding timestep indices $\{t_i\}_{i=0}^{1000}$. For a given $\bar{\alpha}_{\text{target}}$, we use binary search to locate the nearest pair of entries $(\bar{\alpha}_{j-1}, \bar{\alpha}_{j})$ such that $\bar{\alpha}_{j-1} \leq \bar{\alpha}_{\text{target}} \leq \bar{\alpha}_{j}$. We then perform linear interpolation in timestep space:
\begin{equation}
	t^* = t_{j-1} + \frac{\bar{\alpha}_{\text{target}} - \bar{\alpha}_{j-1}}{\bar{\alpha}_{j} - \bar{\alpha}_{j-1}} (t_{j} - t_{j-1})
	\label{eq:alpha_inversion}
\end{equation}
The resulting timestep inversion strategy converts reconstruction uncertainty into an adaptive diffusion timestep through a differentiable noise-level prediction followed by fast lookup-based inversion.

\subsection{Lightweight Student Model Construction}
In the second stage, we aim to lightweight the teacher model obtained in the first stage to derive a more efficient student model, as illustrated in Fig.~\ref{fig:progr_pruned} and Table~\ref{tab:sd21_complexity}. For latent diffusion models such as Stable Diffusion v2.1, the majority of parameters are concentrated in the UNet, while the Variational Autoencoder (VAE) contributes a relatively smaller portion of parameters but incurs notable computational overhead due to its operations are performed in the high-resolution image space. 

As summarized in Table~\ref{tab:sd21_complexity}, the UNet overwhelmingly dominates the model capacity, accounting for approximately $92.02\%$ of the total parameters while contributing only $16\%$ of the overall computational cost. In contrast, the VAE contains merely $7.98\%$ of the parameters but incurs a significantly higher computational burden, accounting for about $84\%$ of the total computation. This discrepancy reveals a pronounced imbalance between parameter distribution and computational cost across different components. Therefore, we apply dedicated lightweight optimization strategies to both the VAE and the UNet, leading to the following three approaches.

\begin{table}[htbp]
	\centering
	\small
	\renewcommand{\arraystretch}{1.2}
	\caption{Proportion of Parameters and Computational Complexity of VAE and UNet in a Single-Step Stable Diffusion Model.}
	\label{tab:sd21_complexity}
	\begin{tabular}{lcc}
		\toprule
		Component & Parameter Ratio & Computation Ratio (MACs) \\
		\midrule
		UNet & $ 92.02\%$ & $ 16\%$ \\
		VAE  & $ 7.98\%$& $ 84\%$ \\
		\bottomrule
	\end{tabular}
\end{table}

\begin{figure}[htb]
	\centering
	\includegraphics[width=\linewidth]{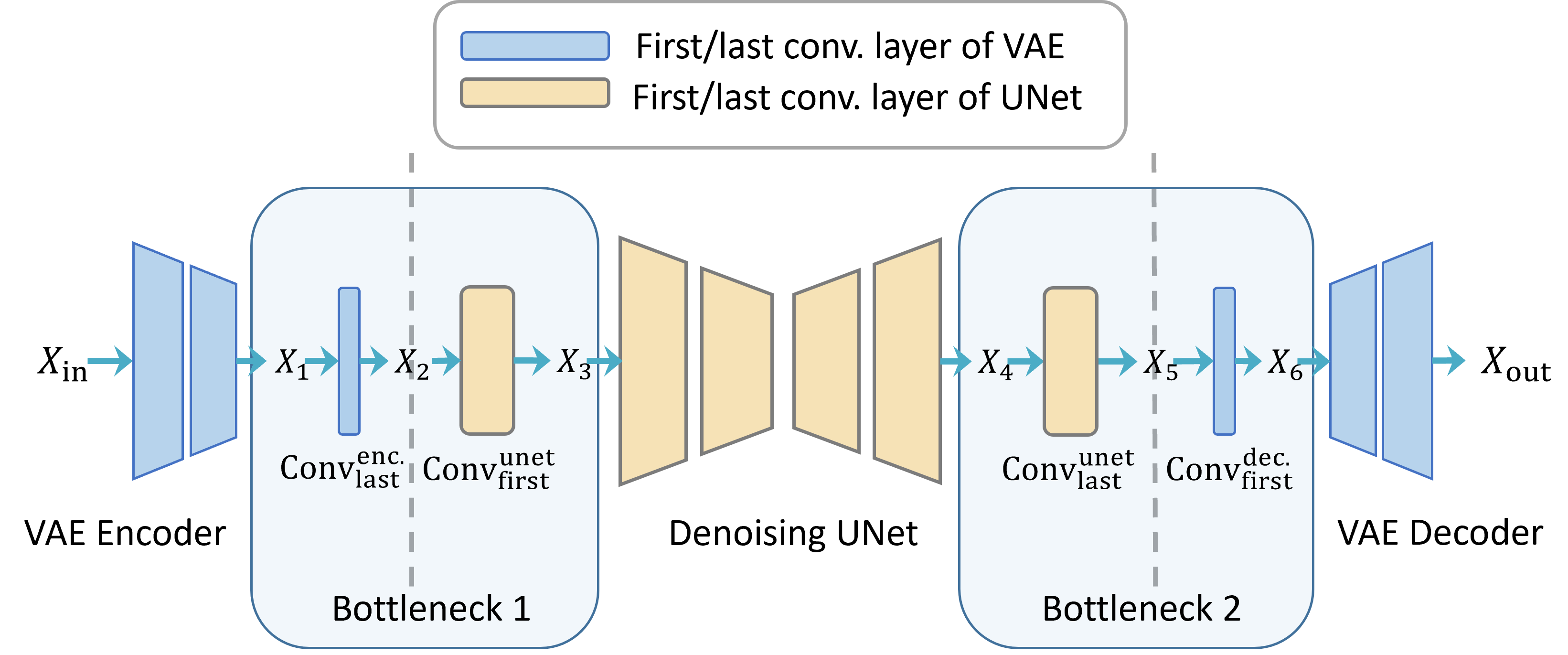}
	\caption{For the bottleneck between the VAE and the UNet, the last convolutional layer of the VAE encoder and the first convolutional layer of the UNet respectively perform dimensionality reduction and expansion on the latent features. These two layers together form a bottleneck structure. During the transformation from $X_1$ to $X_2$, the feature is heavily compressed and cannot be fully recovered in the subsequent mapping from $X_2$ to $X_3$. A similar situation also occurs between the last convolutional layer of the UNet and the first convolutional layer of the VAE decoder.}
	\label{fig:bottleneck}
\end{figure}

\subsubsection{Enhanced Lightweight VAE}
To reduce the additional computational cost and parameter overhead introduced by the VAE, we replace the original VAE in the latent diffusion model with an extremely lightweight VAE model, TAESD \footnote{\url{https://github.com/madebyollin/taesd}}, a compact autoencoder that approximates the functionality of the VAE of Stable Diffusion models while significantly reducing model size and inference cost. Although efficient, TAESD still fails to address the information bottleneck between the UNet and the VAE in the original Stable Diffusion architecture. As illustrated in Fig.~\ref{fig:bottleneck}, the last convolutional layer of the VAE encoder significantly compresses the number of feature channels, while the first convolutional layer of the UNet substantially expands the feature dimensionality:
\begin{equation}
	X_2 = \text{Conv}_{\text{last}}^{\text{enc.}}\left(X_1\right), \quad
	X_3 = \text{Conv}_{\text{first}}^{\text{unet}}\left(X_2\right)
\end{equation}
where $X_1 \in \mathbb{R}^{B \times 64 \times H \times W}$, $X_2 \in \mathbb{R}^{B \times 4 \times H \times W}$, and $X_3 \in \mathbb{R}^{B \times 320 \times H \times W}$ denote the feature representations before compression, after compression, and after expansion, respectively. Here, the change in the three feature channel dimensions indicates that the feature information is heavily compressed when transforming from $X_1$ to $X_2$, which creates an information bottleneck between the VAE encoder and the UNet. Similarly, the same information bottleneck also exists between the last convolutional layer of the UNet and the first convolutional layer of the VAE decoder. 

The presence of such bottleneck layers significantly compresses the edge details of the LR input images, which can adversely affect the subsequent reconstruction process, particularly in terms of fidelity. To address this issue, we propose a convolution fusion strategy, a simple yet effective approach to alleviate the information bottleneck between the VAE and the UNet. Specifically, we fuse the two consecutive convolution operations into a single convolutional transformation. Instead of first compressing the feature channels from $64$ to $4$ and then expanding them to $320$, the proposed strategy directly projects the feature representation from $64$ channels to $320$ channels:
\begin{equation}
	X_3 = \text{Conv}_{\text{fuse}}\left(X_1\right)
	\label{eq:conv_fuse}
\end{equation}
where $\text{Conv}_{\text{fuse}}$ denotes the fused convolution operation. By bypassing the intermediate $4$-channel latent representation $X_2$, the proposed design effectively removes the severe information bottleneck between the VAE encoder and the UNet, allowing richer structural and edge information to be preserved during feature propagation. Similarly, we employ another fused convolution to replace the last convolutional layer of the UNet and the first convolutional layer of the VAE decoder, thereby alleviating the information bottleneck between the UNet and the VAE decoder.

\subsubsection{Module Removal in UNet}
The original UNet in Stable Diffusion is designed for generative tasks and thus contains many components that are redundant for SR. We adopt two strategies to remove these unnecessary modules. First, following AdcSR \cite{chen2025adversarial}, we remove the text encoder and cross-attention (CA) layers, since textual conditioning provides limited guidance for the pruned student network. We also remove time embeddings in the student model. This design is motivated by the different roles of the teacher and student models. The teacher model explicitly uses timestep conditioning to adapt its generative strength according to reconstruction difficulty. In contrast, the student model is trained via distillation to approximate the behavior of the teacher across different timesteps. As a result, the effect of timestep variation is implicitly encoded into the student network parameters. Therefore, explicit timestep conditioning becomes redundant for the student model. Removing the timestep embedding thus simplifies the model and improves efficiency, without sacrificing reconstruction quality.
\begin{figure}[htb]
	\centering
	\includegraphics[width=\linewidth]{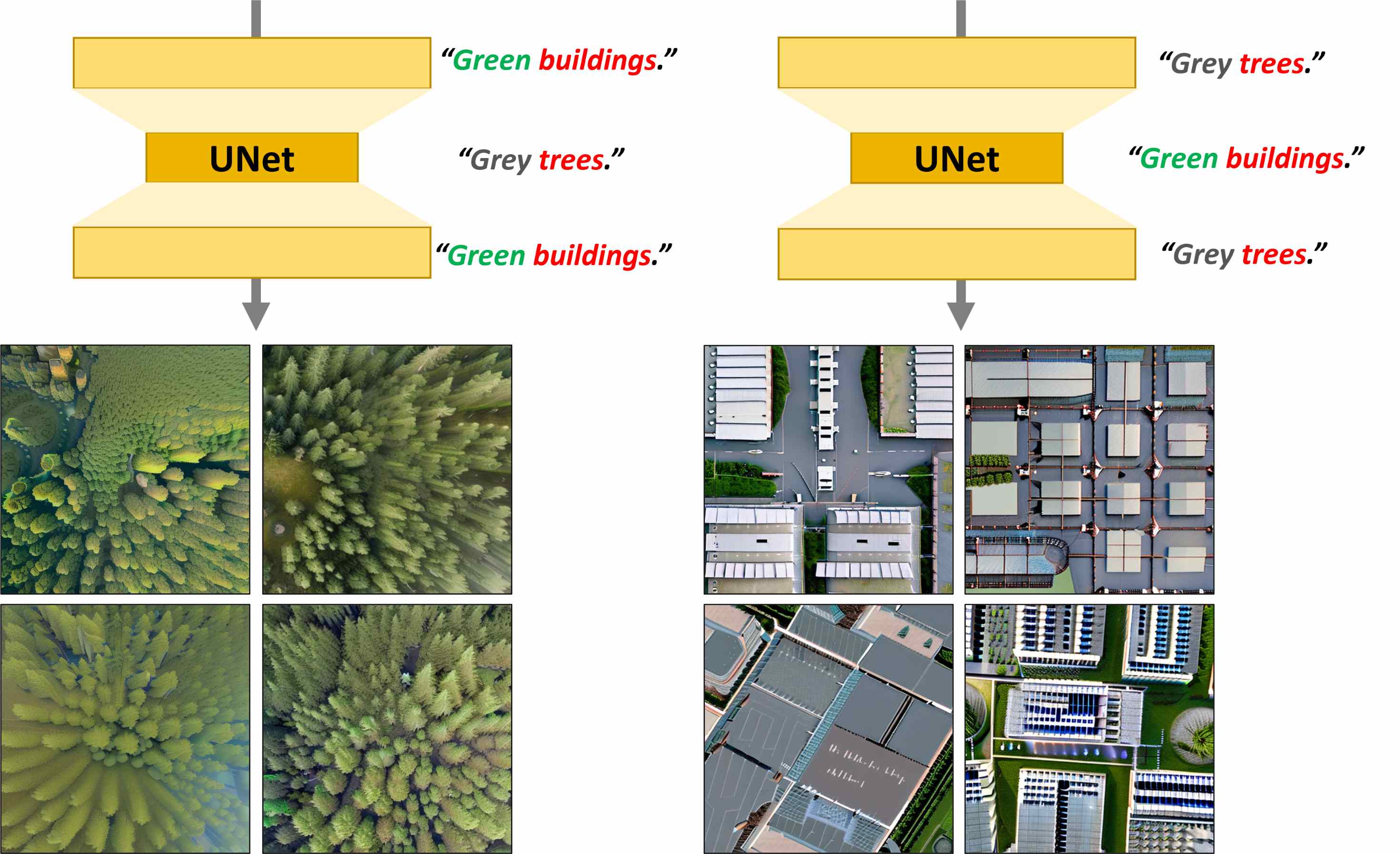}
	\caption{We provide different textual prompts for the cross-attention layers at varying depths of the UNet. In the figure, we have omitted the prompt "a satellite image with a top-down perspective" and only retained keywords related to style and content, such as "green/grey" and "trees/buildings." It can be observed that the details of the generated image is determined by the shallow layers of the network, while the global content is governed by the deeper layers.}
	\label{fig:deep_shallow}
\end{figure}

\begin{figure}[htb]
	\centering
	\includegraphics[width=\linewidth]{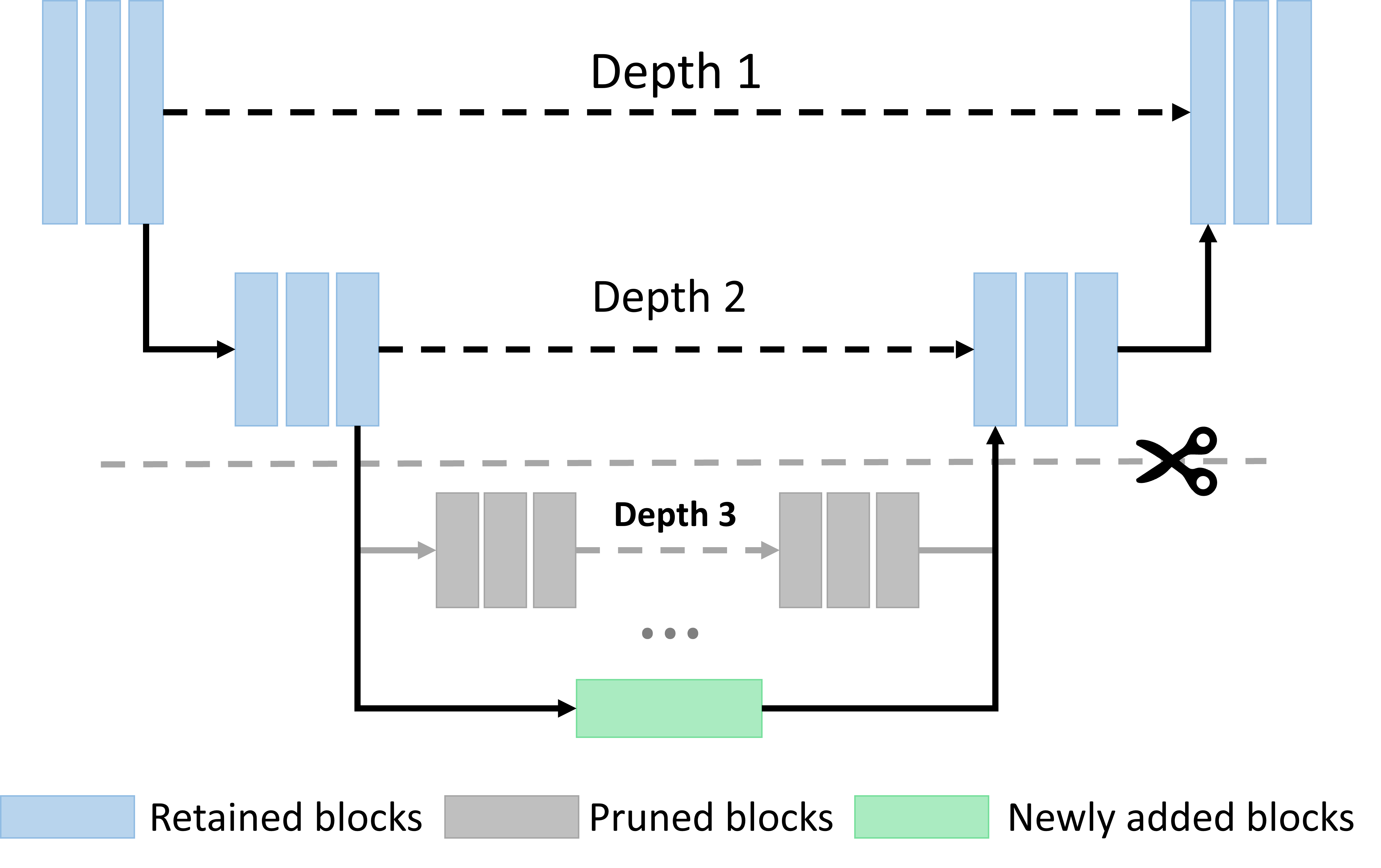}
	\caption{Our proposed semantic-aware pruning strategy entirely removes the two deepest layers of the UNet, while all shallow layers are fully retained. To ensure structural consistency, the removed deep modules are replaced with a simple ResBlock to preserve the skip connections between the encoder and decoder of the UNet.}
	\label{fig:unet_prun}
\end{figure}
Next, we consider how to further prune the remaining modules. AdcSR applies uniform channel pruning to UNet modules of different depths. However, previous studies \cite{voynov2023p+, li2025megasr} have shown that deep modules in UNet tend to reconstruct structure and semantics, while shallow modules are responsible for reconstructing details and appearance. As shown in Fig. \ref{fig:deep_shallow}, we provide different prompts for the CA layers at different depths. From the final generated images, it is evident that the semantics of the results follow the deep prompts, while the appearance adheres to the shallow prompts. However, for SR tasks, input LR images already provide sufficient semantic information, making the deep modules in UNet redundant for SR. Therefore, unlike AdcSR, which applies uniform channel pruning across UNet modules at different depths, we propose a semantic-aware pruning strategy that completely removes the deeper UNet blocks and replaces them with a single ResBlock to maintain structural integrity, as illustrated in Fig. \ref{fig:unet_prun}. Since deeper modules typically contain a larger number of channels, pruning them can significantly reduce the model parameters while minimizing the impact on the reconstruction capability of the student model.

\begin{figure}[htb]
	\centering
	\includegraphics[width=\linewidth]{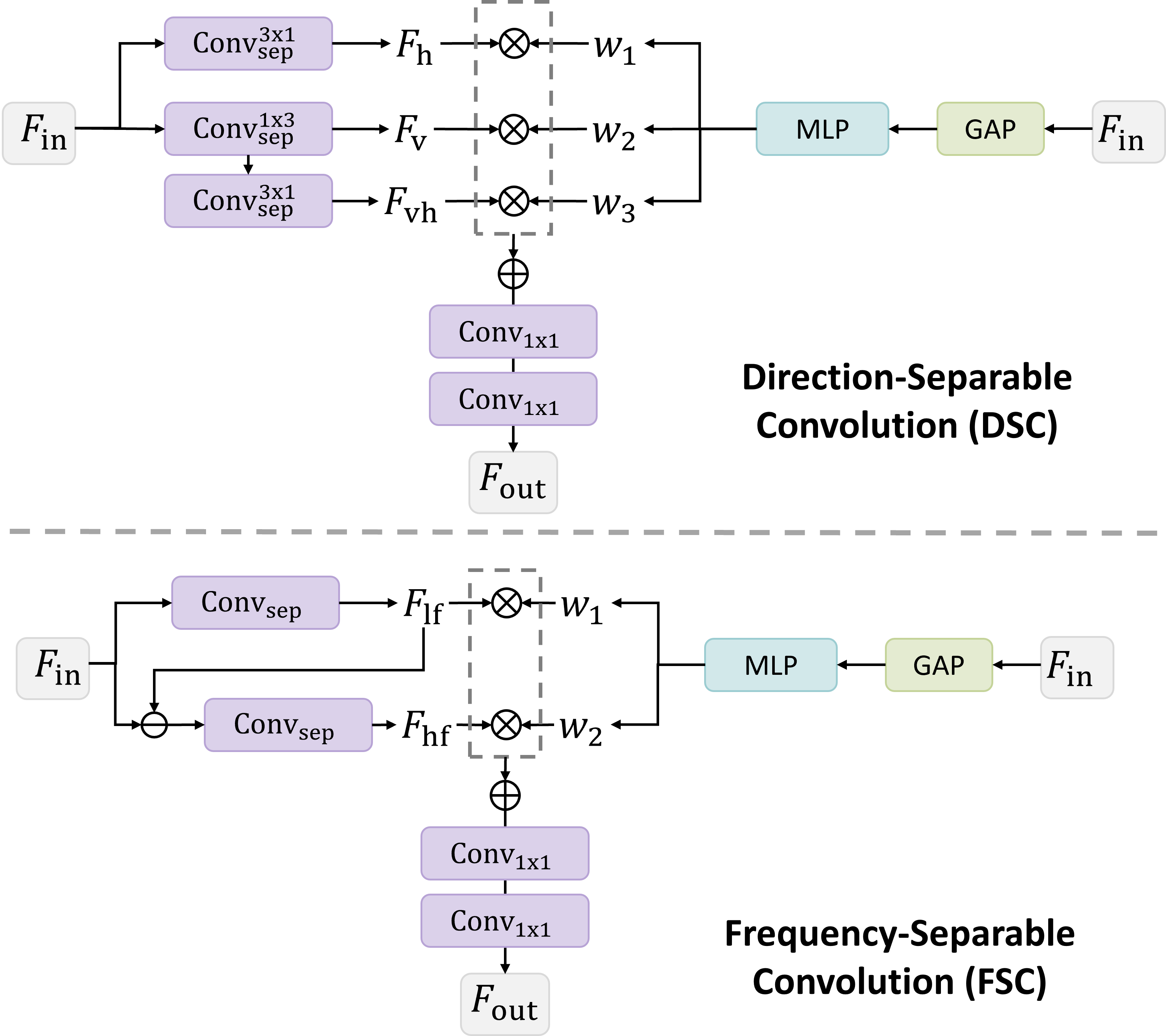}
	\caption{The architecture diagrams of direction-separable convolution (DSC) and frequency-separable convolution (FSC). Both modules follow a similar pipeline: they first employ specialized convolutions to extract multi-source features, then perform weighted fusion over these features, and finally use two $1 \times 1$ convolutions to complete channel transformation.}
	\label{fig:seperable_conv}
\end{figure}

\begin{figure}[htb]
	\centering
	\includegraphics[width=\linewidth]{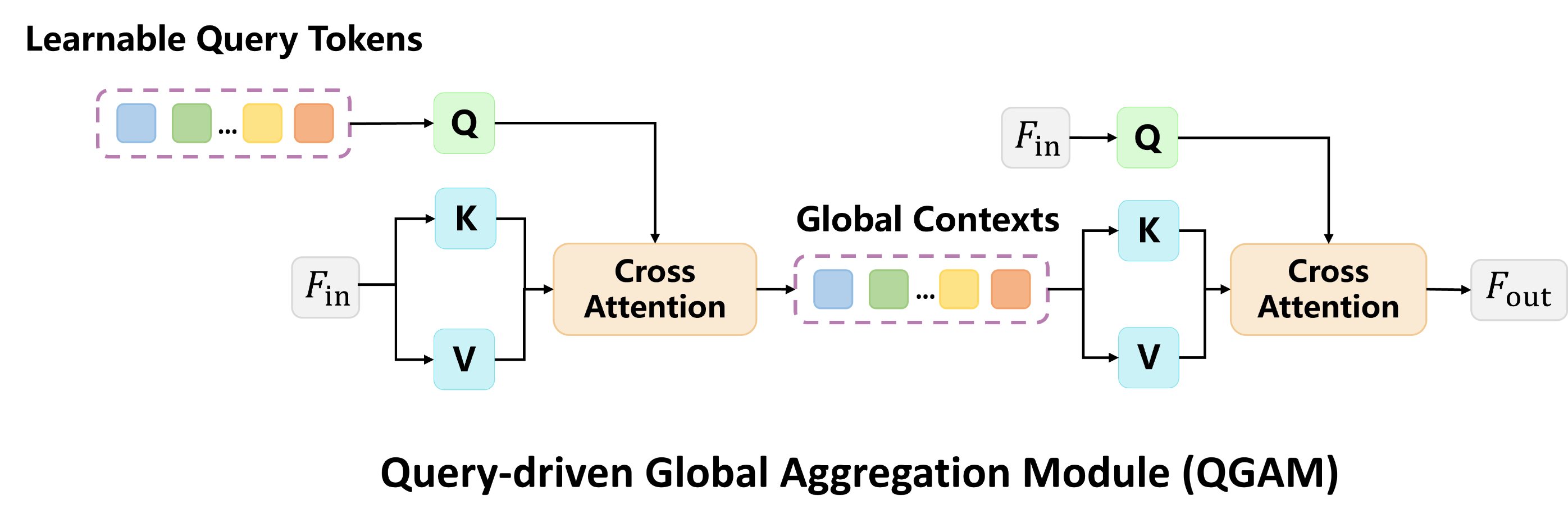}
	\caption{The architecture diagram of the query-driven global aggregation module (QGAM). A set of learnable query tokens is first introduced, which interact with the input features via cross-attention to generate global contextual representations. Subsequently, the global context is propagated back to the input features through another cross-attention operation.}
	\label{fig:learn_query}
\end{figure}

\subsubsection{Module Replacement in UNet}
To further improve efficiency, we replace the remaining UNet modules with lightweight components tailored to remote sensing imagery. It should be emphasized that these modules are designed primarily as lightweight substitutes for standard operations, rather than as performance-enhancing modules. Their objective is to reduce model size and computational cost while preserving reconstruction capability. Remote sensing images typically exhibit large homogeneous regions with limited low-frequency variation, while critical information is concentrated in sparse high-frequency structures such as edges and object boundaries. To exploit this property, we introduce frequency-separable convolution (FSC), which decouples low and high-frequency components, allowing the network to focus computational resources on informative details. In addition, remote sensing scenes often contain strong directional patterns, such as roads, rivers, and building layouts. To capture such anisotropic structures efficiently, we propose direction-separable convolution (DSC), which decomposes convolution operations into directional components. As illustrated in Fig. \ref{fig:seperable_conv}, both modules follow a similar pipeline: they first employ specialized convolutions to extract features from different aspects, then perform weighted fusion over these features, and finally use two $1 \times 1$ convolutions to complete channel transformation. For DSC, we use horizontal and vertical convolutions to extract features along horizontal, vertical, and diagonal directions. For FSC, we employ two convolutions to separately capture low-frequency information from the input features and high-frequency information from the residual features. It is worth noting that, except for the final two $1 \times 1$ convolutions used for channel transformation, all other convolutions are depthwise separable convolutions, which significantly reduce the number of parameters compared to the standard $3 \times 3$ convolutions in the original UNet.

Furthermore, due to the large spatial extent of remote sensing imagery, long-range dependencies are essential but computationally expensive to model with standard self-attention. Remote sensing imagery often contains a large number of repetitive and limited land-cover categories \cite{guo2024skysense}, which can be effectively represented by a set of learnable queries. Therefore, we introduce a query-driven global aggregation module (QGAM), which approximates global context modeling through learnable queries, reducing complexity from quadratic to linear while preserving global interactions.  As illustrated in Fig. \ref{fig:learn_query}, QGAM can be formulated as two cross-attention operations. Given input features $F_{\text{in}} \in \mathbb{R}^{N \times d}$ and learnable queries $Q \in \mathbb{R}^{M \times d}$ ($M \ll N$), we first aggregate global context tokens:
\begin{equation}
	G = \mathrm{CA}(Q, F_{\text{in}}) = \mathrm{Softmax}\!\left(\frac{(QW_q)(F_{\text{in}}W_k)^\top}{\sqrt{d}}\right)F_{\text{in}}W_v,
\end{equation}
and then propagate the global context back to feature tokens:
\begin{equation}
	F_{\text{out}} = \mathrm{CA}(F_{\text{in}}, G) = \mathrm{Softmax}\!\left(\frac{(F_{\text{in}}W'_q)(GW'_k)^\top}{\sqrt{d'}}\right)GW'_v.
\end{equation}
In this way, the computational complexity is reduced from $\mathcal{O}(N^2)$ to $\mathcal{O}(NM)$ while preserving global interactions among input features.

\subsection{Knowledge Distillation}
To minimize the performance gap between the lightweight student model and the teacher model, we employ knowledge distillation techniques to transfer the generative capabilities from the teacher to the student. Traditional feature distillation commonly relies on MSE loss, which enforces point-wise alignment between corresponding teacher and student features. However, point-wise alignment alone may be insufficient to capture distribution-level discrepancies between feature representations, potentially leading to suboptimal distillation outcomes.

To address this limitation, we introduce the Maximum Mean Discrepancy (MMD) \cite{gretton2012kernel} loss in addition to the MSE loss to encourage distribution-level alignment between the teacher and student features. MMD is a kernel-based metric that measures the difference between two probability distributions by computing the maximum discrepancy over a class of functions. Specifically, we use the radial basis function (RBF) kernel to compute MMD:

\begin{equation}
	\begin{aligned}
		\text{MMD}^2(P, Q) =&\; \mathbb{E}_{x,x' \sim P}[k(x, x')] + \mathbb{E}_{y,y' \sim Q}[k(y, y')] \\
		&- 2\mathbb{E}_{x \sim P, y \sim Q}[k(x, y)]
	\end{aligned}
	\label{eq:mmd}
\end{equation}
where $(x, x')$ and $(y, y')$ denote samples drawn from distributions $P$ and $Q$, respectively, and $k(\cdot, \cdot)$ is the RBF kernel defined as $k(x, y) = \exp\left(-\frac{|x - y|^2}{2\sigma^2}\right)$ with bandwidth $\sigma$.

In our distillation process, for each corresponding layer between the teacher and student networks, we first apply a feature adapter $\phi$ (a lightweight convolutional module) to align the channel dimensions of the student features with those of the teacher. We then compute the MSE loss directly on the aligned features:

\begin{equation}
	\mathcal{L}_{\text{MSE-Dis}} = \frac{1}{N} \sum_{i=1}^N \| \phi\left(f_{\text{stu}}^i\right) - f_{\text{tea}}^i \|^2
	\label{eq:mse_loss}
\end{equation}
where $f_{\text{stu}}^i$ and $f_{\text{tea}}^i$ are the feature maps from the student and teacher networks at layer $i$, and $N$ is the number of the remained UNet layers.

Additionally, we flatten the feature maps into sample sets and compute the MMD loss to capture distributional differences:

\begin{equation}
	\mathcal{L}_{\text{MMD-Dis}} = \text{MMD}^2(P_{\text{stu}}, P_{\text{tea}})
	\label{eq:mmd_loss}
\end{equation}

The total feature distillation loss for each layer is a weighted combination of MSE and MMD:

\begin{equation}
	\mathcal{L}_{\text{Dis.}} = \mathcal{L}_{\text{MSE-Dis}} + \lambda \mathcal{L}_{\text{MMD-Dis}}
	\label{eq:total_feat_loss}
\end{equation}
where $\lambda = 0.1$ is a hyperparameter balancing the two losses. This multi-scale feature distillation strategy ensures that the student model not only matches the point-wise statistics but also aligns the distribution-level discrepancies of the intermediate features, leading to improved visual quality and reconstruction fidelity.

\subsection{Training Scheme}

As illustrated in Fig. \ref{fig:progr_pruned}, our method consists of two main training stages: teacher model pretraining and knowledge distillation from teacher to student.

\subsubsection{Stage 1: Teacher Model Pretraining}

In training the teacher model, we freeze the pre-trained uncertainty estimator $\mathrm{UE}_{\phi}$ and use LoRA \cite{hu2022lora} to fine-tune the linear and convolutional layers in the UNet and VAE-Decoder. For the loss function, we employ a weighted combination of reconstruction and regularization objectives. The reconstruction loss is defined as:

\begin{equation}
	\mathcal{L}_{\text{Rec.}} = \mathcal{L}_{\text{L1}} + \lambda_{\text{LPIPS}} \mathcal{L}_{\text{LPIPS}} + \lambda_{\text{GAN}} \mathcal{L}_{\text{GAN}}
	\label{eq:teacher_rec_loss}
\end{equation}

where $\mathcal{L}_{\text{L1}} = \| \hat{x} - x_{\text{gt}} \|_1$, $\mathcal{L}_{\text{LPIPS}}$ is the perceptual loss, and $\mathcal{L}_{\text{GAN}}$ is the adversarial loss. We initialize the discriminator parameters using the pre-trained visual encoder DINO \cite{caron2021emerging} to leverage its rich semantic representation capabilities. In addition to reconstruction fidelity, we incorporate a total variation (TV) regularization term on the predicted noise level $\bar{\alpha}_{\text{target}}$ to encourage spatial smoothness:

\begin{equation}
	\mathcal{L}_{\text{TV}} = \| \nabla_h \bar{\alpha}_{\text{target}} \|_2^2 + \| \nabla_w \bar{\alpha}_{\text{target}} \|_2^2
	\label{eq:tv_loss}
\end{equation}

where $\nabla_h$ and $\nabla_w$ denote spatial gradients along height and width dimensions, respectively. This regularization prevents abrupt transitions in the predicted timesteps and promotes consistent adaptive generative guidance across neighboring spatial regions. The total teacher training loss combines both terms:

\begin{equation}
	\mathcal{L}_{\text{Teacher}} = \mathcal{L}_{\text{Rec.}} + \lambda_{\text{TV}} \mathcal{L}_{\text{TV}}
	\label{eq:teacher_loss}
\end{equation}

\subsubsection{Stage 2: Knowledge Distillation from Teacher to Student}

After completing the teacher model training, we use the teacher model as supervision to train the student model. The retained modules in the student model are initialized with pre-trained Stable Diffusion-2.1 parameters, while any newly added modules are initialized with random parameters. All parameters of the student model (including the pruned UNet and enhanced lightweight VAE) participate in training via full fine-tuning. For the loss function, in addition to the same reconstruction losses Eq. \ref{eq:teacher_rec_loss} used in teacher model training, we add the distillation loss Eq. \ref{eq:total_feat_loss} to ensure that the generative priors of the teacher model can adequately guide the training of the student model:

\begin{equation}
	\mathcal{L}_{\text{Student}} = \mathcal{L}_{\text{Rec.}} + \mathcal{L}_{\text{Dis.}}
	\label{eq:student_loss}
\end{equation}
\section{Experiments and discussion}
\subsection{Dataset}
We use all 10,000 images from the AID \cite{xia2017aid} dataset as the training set, and sample 630 images from NWPU-RESISC45 \cite{cheng2017remote}, 500 from DOTA \cite{xia2018dota}, and 500 from DIOR \cite{li2020object} as the test set. All images retain their original sizes without any cropping operations. To construct LR images, we follow the degradation process of Real-ESRGAN \cite{wang2021real} and consider two magnification factors, $4 \times$ and $8 \times$, thereby enabling a comprehensive evaluation of the reconstruction performance of different methods under varying scaling factors.

Beyond synthetic degradation, we also evaluate on a real-world remote sensing dataset Real-RefRSSRD. \cite{wang2026controllable}, which provides paired images at different spatial resolutions from actual remote sensing systems. Specifically, the LR images are from  Sentinel-2 with a ground sampling distance (GSD) of 10~m, and the HR images are from NAIP (National Agriculture Imagery Program) data with a GSD of 1~m. This dataset is particularly challenging because it involves real degradation processes and large-scale spatial variations inherent to actual Earth observation platforms, rather than synthetic degradation commonly used in standard benchmarks.

\subsection{Implementation Details}
We conduct the two-stage training on 2 RTX 3090 GPUs. In training the teacher model, we use the AdamW optimizer with a learning rate of 2e-4, for 25,000 iterations with a batch size of 2. The LoRA ranks for the UNet and VAE-Decoder are set to 32 and 16, respectively. During training, input images are randomly cropped to $512 \times 512$. The weights $\lambda_{\text{LPIPS}}$ and $\lambda_{\text{GAN}}$ are set to 2.5 and 0.25, respectively. In training the student model, we use the Adam optimizer with a learning rate of 5e-5, for 100,000 iterations with a batch size of 16. The deepest two layers of the student model's UNet are completely removed and replaced with a simple ResBlock. For the remaining UNet modules, we replace them with our proposed lightweight alternative modules. During training, input images are also randomly cropped to $512 \times 512$.

\subsection{Evaluation Metrics}
To comprehensively evaluate different SR methods, we adopt both full-reference and no-reference image quality metrics. For pixel-level fidelity, we employ peak signal-to-noise ratio (PSNR) and structural similarity index (SSIM), both computed on the Y channel of the YCbCr color space. For perceptual quality assessment, we utilize learned perceptual image patch similarity (LPIPS) and deep image structure and texture similarity (DISTS).

Furthermore, two no-reference metrics, MUSIQ \cite{ke2021musiq} and MANIQA \cite{yang2022maniqa}, are introduced to provide a more holistic evaluation of overall performance. In terms of efficiency, we consider inference time and model size (number of parameters). The inference time is measured in seconds for a $512 \times 512$ image on a single RTX 3090 GPU.

\subsection{Comparison With State-of-the-Art Methods}
To comprehensively evaluate the performance of our proposed method, we compare it against eight state-of-the-art SR methods, including two recent regression-based Transformer methods (TTST \cite{xiao2024ttst} and TransENet \cite{lei2021transformer}), two classical GAN-based methods (ESRGAN \cite{wang2018esrgan} and SwinIR-GAN \cite{liang2021swinir}), three recent diffusion-based methods (EDiffSR \cite{xiao2023ediffsr}, DiffBIR \cite{lin2024diffbir}, and S3Diff \cite{zhang2024degradation}), and one closely related lightweight diffusion-based SR method (AdcSR \cite{chen2025adversarial}). For a fair comparison, all competing methods are retrained on our dataset using their original code and configurations.

\begin{table*}[htb]
	\centering
	\caption{Quantitative comparison of different methods on the 4$\times$  test datasets. The best, second-best, and third-best results are highlighted in \best{bold red}, \second{underlined blue}, and \third{italic green}, respectively.}
	\label{tab:comparison}
	\resizebox{\textwidth}{!}{
	\begin{tabular}{@{\extracolsep{\fill}}llcccccccc@{}}
		\toprule
		\multirow{2}{*}{Category} & \multirow{2}{*}{Method} & \multicolumn{6}{c}{Image Quality Metrics} & \multicolumn{2}{c}{Efficiency Metrics} \\
		\cmidrule(lr){3-8} \cmidrule(lr){9-10}
		& & PSNR $\uparrow$ & SSIM $\uparrow$ & LPIPS $\downarrow$ & DISTS $\downarrow$ & MANIQA $\uparrow$ & MUSIQ $\uparrow$ & Time (s) $\downarrow$ & Param. (M) $\downarrow$ \\
		\midrule
		
		\multirow{2}{*}{\textbf{Regression-based}} & TTST
		& \second{26.05}
		& \best{0.6663}
		& 0.5309
		& 0.2825
		& 0.2397
		& 38.55
		& 0.218
		& \third{18.37} \\
		
		& TransENet
		& \best{26.17}
		& \second{0.6651}
		& 0.5240
		& 0.2826
		& 0.2381
		& 37.50
		& \third{0.077}
		& 35.09 \\
		
		\midrule
		
		\multirow{2}{*}{\textbf{GAN-based}} & ESRGAN
		& 24.71
		& 0.5919
		& 0.3888
		& \third{0.2074}
		& 0.2442
		& 48.18
		& \second{0.044}
		& \second{16.72} \\
		
		& SwinIR-GAN
		& \third{25.18}
		& \third{0.6100}
		& \third{0.3753}
		& 0.2189
		& 0.2416
		& 45.34
		& 0.132
		& \best{11.46} \\
		
		\midrule
		
		\multirow{5}{*}{\textbf{Diffusion-based}} & EDiffSR
		& 23.19
		& 0.4896
		& 0.4512
		& 0.2799
		& 0.1839
		& 42.55
		& 6.745
		& 30.37 \\
		
		& DiffBIR
		& 21.55
		& 0.3986
		& 0.4738
		& 0.2489
		& \best{0.4523}
		& \best{60.86}
		& 8.637
		& 1717 \\
		
		& S3Diff
		& 23.18
		& 0.5112
		& \second{0.3511}
		& \second{0.1941}
		& \second{0.2855}
		& \second{51.02}
		& 0.736
		& 1327 \\
		
		& AdcSR
		& 24.64
		& 0.5918
		& 0.3935
		& 0.2256
		& 0.2717
		& 44.94
		& 0.088
		& 456.1 \\
		
		& Ours
		& 24.09
		& 0.5642
		& \best{0.3443}
		& \best{0.1819}
		& \third{0.2837}
		& \third{49.84}
		& \best{0.037}
		& 58.05 \\
		
		\bottomrule
	\end{tabular}}
\end{table*}

\subsubsection{Quantitative Comparison}
Table \ref{tab:comparison} presents the quantitative comparison results on the test datasets for $4\times$ SR. Our method achieves competitive performance across various metrics while maintaining high inference efficiency.

First, in fidelity-oriented pixel metrics such as PSNR and SSIM, regression-based methods (TTST and TransENet) achieve the best scores, as they are directly optimized toward pixel-wise losses. However, they tend to produce over-smoothed results, leading to inferior perceptual performance. Second, for perceptual full-reference metrics such as LPIPS and DISTS, our method achieves the best performance. This advantage arises from its ability to leverage the intrinsic priors of pretrained diffusion models, together with a compact architecture that facilitates efficient adaptation to the distribution of remote sensing imagery. Finally, multi-step diffusion models such as DiffBIR, as well as unpruned diffusion methods like S3Diff, exhibit the strongest generative capacity and therefore achieve superior results on no-reference metrics such as MANIQA and MUSIQ. However, they often introduce hallucinated structures, which degrade performance on full-reference metrics. Overall, our method not only preserves superior perceptual fidelity but also effectively exploits the generative capability provided by pretrained diffusion priors, achieving a well-balanced reconstruction performance across all metrics.

In terms of inference efficiency, our method demonstrates substantial improvements over other diffusion-based approaches. For example, compared with the multi-step diffusion model DiffBIR and the single-step diffusion model S3Diff, our method achieves more than $200\times$ and $20\times$ speedups in inference time, respectively. Compared with the lightweight diffusion model AdcSR, our approach further reduces the number of parameters by nearly $10\times$, thereby significantly lowering the hardware requirements for deployment. Moreover, it is worth noting that our method even surpasses ESRGAN, a classic GAN-based approach, in inference speed. This is mainly because, benefiting from the unique design of latent diffusion models, most of the computation in our method is performed in a latent space with an $8\times$ downsampling factor.

\begin{figure*}
	\centering
	\includegraphics[width=0.9\linewidth]{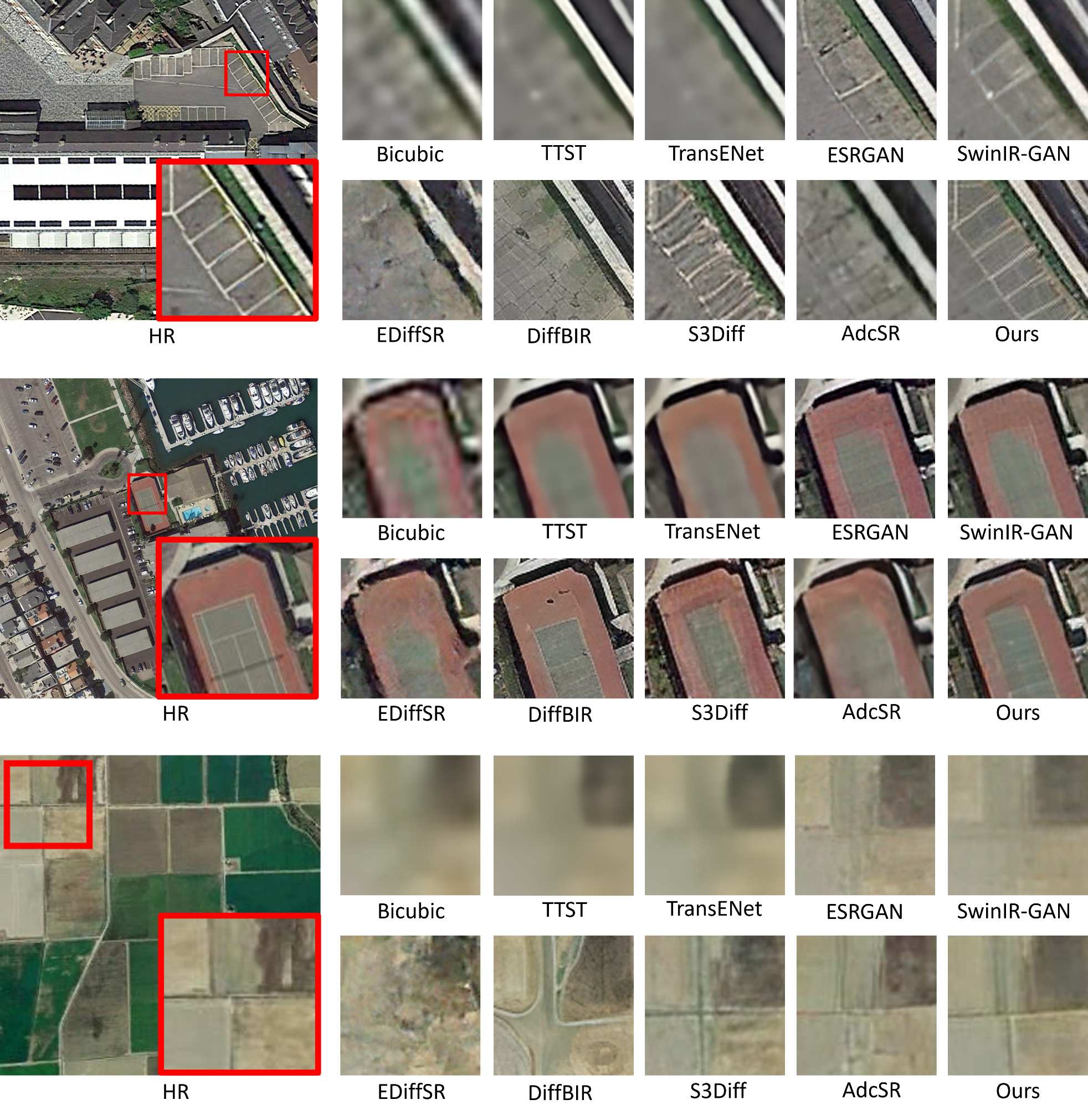}
	\caption{Visual comparison of $4\times$ SR results of different methods on the DIOR, DOTA, and NWPU-RESISC45 datasets. The results demonstrate that our method is capable of reconstructing more accurate structures and richer edge details.}
	\label{fig:compare_4x}
\end{figure*}

\subsubsection{Qualitative Comparison}
In Fig. \ref{fig:compare_4x}, we present a visual comparison of $4\times$ SR results produced by different methods. The three groups of images, from top to bottom, are sampled from the test sets of the DIOR, DOTA, and NWPU-RESISC45 datasets, respectively. It can be observed that regression-based methods such as TTST and TransENet can only generate overly smooth results, failing to faithfully reconstruct the complex structures and rich edge details in remote sensing imagery. In contrast, ESRGAN and SwinIR-GAN incorporate perceptual losses and adversarial training strategies, leading to significantly improved visual quality compared to the former methods. However, due to the highly complex degradation process in real-world remote sensing images, simple adversarial training is insufficient to introduce strong priors. As a result, although the outputs appear sharper, these methods often produce distorted structures (e.g., the ESRGAN results in the first and second groups).

By comparison, diffusion-based methods such as DiffBIR and S3Diff effectively leverage pretrained diffusion priors and are able to generate richer details. Nevertheless, they tend to suffer from severe hallucination effects. For instance, in the third group, DiffBIR and S3Diff generate curved and cross-shaped farmland boundaries, respectively, which are inconsistent with real-world land surface structures. As a lightweight diffusion model, AdcSR produces overly smooth results. On the one hand, it heavily relies on a pretrained teacher model and is constrained only by an L2 loss in the latent space, which can lead to suboptimal performance when the teacher model is not sufficiently strong. On the other hand, AdcSR adopts uniform channel pruning across the all layers of UNet without fully considering the characteristics of the SR task. In contrast, our method, while remaining highly lightweight, is still capable of reconstructing competitive structures and fine details, thereby providing a more balanced trade-off between efficiency and reconstruction quality.

\subsubsection{Results of $8\times$ SR}
\begin{figure}[htb]
	\centering
	\includegraphics[width=\linewidth]{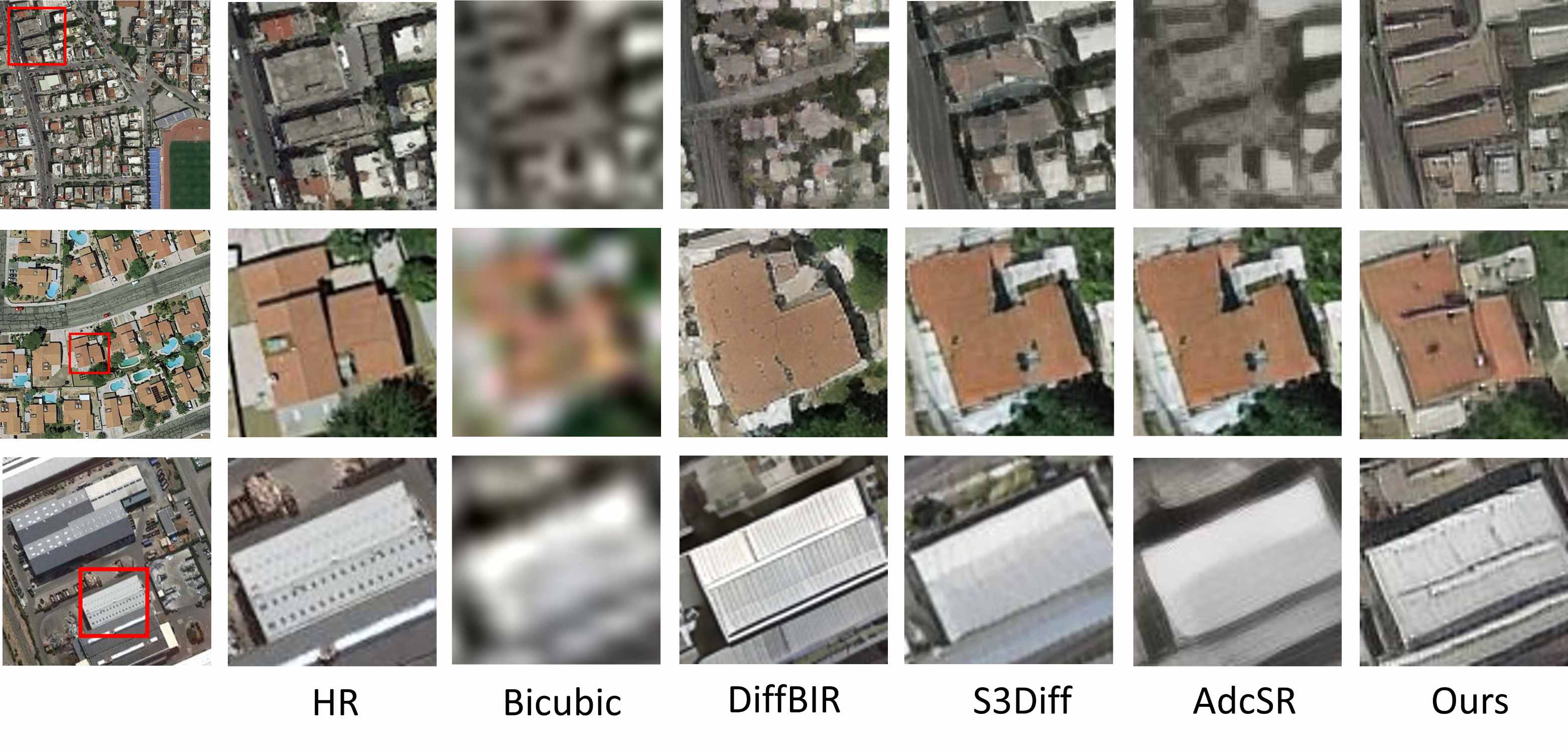}
	\caption{Visual comparison of $8\times$ SR results of different methods on the DIOR, DOTA, and NWPU-RESISC45 datasets.}
	\label{fig:compare_8x}
\end{figure}

Table \ref{tab:comparison-8x} presents the quantitative comparison of different methods on the $8\times$ test set. Since pixel-level metrics (PSNR and SSIM) become less informative at higher scaling factors \cite{zhang2020ntire}, we primarily adopt four perceptual metrics (LPIPS, DISTS, MANIQA and MUSIQ) to evaluate reconstruction quality. It can be observed that our method still maintains satisfactory performance under the more challenging $8\times$ SR setting. However, compared with the $4\times$ case, unpruned models such as S3Diff and DiffBIR demonstrate stronger performance than our approach. This is because, under more severe degradation, the pretrained diffusion priors play a more critical role in improving reconstruction quality. In Fig. \ref{fig:compare_8x}, we present a visual comparison between our method and other diffusion-based approaches. The results show that under the challenging $8\times$ SR setting, DiffBIR suffers from severe hallucination artifacts, producing distorted and unrealistic structures. In contrast, S3Diff, as a single-step diffusion model, enables direct supervision in the pixel space, leading to more realistic and faithful details. Similar to the observations in the $4\times$ setting, AdcSR exhibits noticeable blurring in its reconstructions. By comparison, our method achieves reconstruction quality comparable to S3Diff while significantly improving inference efficiency, demonstrating both its effectiveness and efficiency.
\begin{table}[htb]
	\centering
	\caption{Quantitative comparison of different methods on the 8$\times$ test datasets. The best, second-best, and third-best results are highlighted in \best{bold red}, \second{underlined blue}, and \third{italic green}, respectively.}
	\label{tab:comparison-8x}
	
	\resizebox{\columnwidth}{!}{
		\begin{tabular}{llcccc}
			\toprule
			Category & Method & LPIPS $\downarrow$ & DISTS $\downarrow$ & MANIQA $\uparrow$ & MUSIQ $\uparrow$ \\
			\midrule
			
			\multirow{2}{*}{\textbf{Regression-based}} 
			& TTST 
			& 0.6387 & 0.3430 & 0.1968 & 30.74 \\
			& TransENet 
			& 0.6240 & 0.3399 & 0.1979 & 31.21 \\
			
			\midrule
			
			\multirow{2}{*}{\textbf{GAN-based}} 
			& ESRGAN 
			& 0.4901 & \third{0.2715} & 0.2526 & 44.00 \\
			& SwinIR-GAN 
			& \third{0.4865} & 0.2718 & 0.2199 & 39.94 \\
			
			\midrule
			
			\multirow{5}{*}{\textbf{Diffusion-based}} 
			& EDiffSR 
			& 0.5117 & 0.3118 & 0.1842 & 43.47 \\
			& DiffBIR 
			& 0.5270 & 0.2764 & \best{0.4524} & \best{60.34} \\
			& S3Diff 
			& \best{0.3995} & \best{0.2172} & \third{0.2832} & \third{52.79} \\
			& AdcSR 
			& 0.5026 & 0.3113 & 0.2535 & 39.42 \\
			& Ours 
			& \second{0.4423} & \second{0.2461} & \second{0.3043} & \second{55.32} \\
			
			\bottomrule
		\end{tabular}
	}
\end{table}

\subsubsection{Results of Real-World Dataset}
To further validate effectiveness in real-world scenarios, we evaluate SlimDiffSR on the Real-RefRSSRD dataset \cite{wang2026controllable}. It is important to note that we do not use the cross-sensor paired data provided by Real-RefRSSRD for training. Therefore, the model is not expected to generate SR outputs with NAIP-specific radiometric characteristics. NAIP images are used only as visual references rather than strict radiometric supervision targets.

\begin{figure*}[htb]
	\centering
	\includegraphics[width=\linewidth]{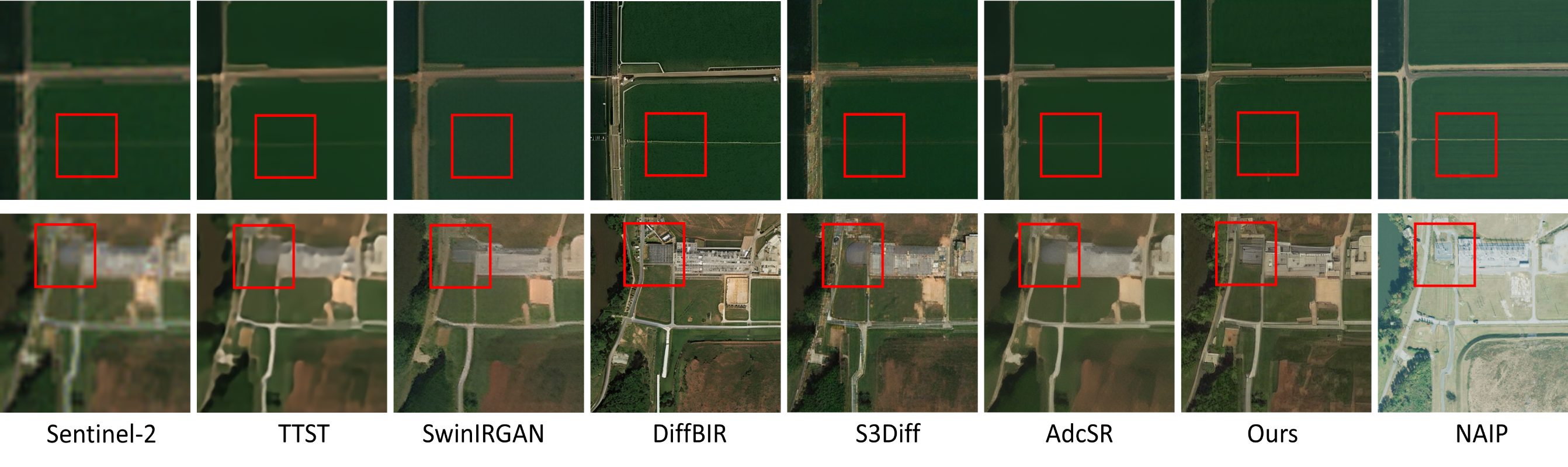}
	\caption{Visual comparison of $8\times$ SR results produced by different methods on the real-world Sentinel-2 SR task. Due to space limitations, for non-diffusion-based approaches, we only present two representative methods, TTST and SwinIRGAN. Compared with other methods, our approach generates more faithful reconstructions with better preservation of spatial structures and fine details.}
	\label{fig:compare_8x_real}
\end{figure*}

Table \ref{tab:comparison-8x-sentinel} presents the quantitative comparison using NIQE, CLIPIQA, MANIQA, and MUSIQ metrics. DiffBIR achieves the best performance by leveraging multi-step denoising and strong priors. Our method achieves second-best results on most metrics while maintaining significant efficiency advantages. Qualitatively, as shown in Fig. \ref{fig:compare_8x_real}, our lightweight model balances reconstruction fidelity and efficiency effectively. While DiffBIR produces visually sharp results, it suffers from hallucination artifacts. In contrast, our method generates faithful spatial structures, preserved boundaries, and rich details comparable to S3Diff while achieving substantially faster inference. These results indicate that the proposed lightweight student is not merely overfitted to the synthetic degradation pipeline, but can generalize to real-world remote sensing degradations while maintaining stable structural reconstruction.
\begin{table}[htb]
	\centering
	\caption{Quantitative comparison of different methods on the $8\times$ Sentinel-2 test set. The best, second-best, and third-best results are highlighted in \best{bold red}, \second{underlined blue}, and \third{italic green}, respectively.}
	\label{tab:comparison-8x-sentinel}
	
	\resizebox{\columnwidth}{!}{
		\begin{tabular}{llcccc}
			\toprule
			Category & Method & NIQE$\downarrow$ & CLIPIQA$\uparrow$ & MANIQA$\uparrow$ & MUSIQ$\uparrow$  \\
			\midrule
			
			\multirow{2}{*}{\textbf{Regression-based}}
			& TTST
			& 10.34
			& 0.1929
			& 0.2076
			& 22.33 \\
			
			& TransENet
			& 9.442
			& 0.1742
			& 0.1993
			& 22.67 \\
			
			\midrule
			
			\multirow{2}{*}{\textbf{GAN-based}}
			& ESRGAN
			& \third{4.229}
			& \third{0.5092}
			& \third{0.2428}
			& 37.84 \\
			
			& SwinIR-GAN
			& 5.365
			& 0.2995
			& 0.1940
			& 28.62 \\
			
			\midrule
			
			\multirow{5}{*}{\textbf{Diffusion-based}}
			& EDiffSR
			& \best{3.651}
			& 0.3567
			& 0.1840
			& \third{39.69} \\
			
			& DiffBIR
			& \second{4.059}
			& \best{0.7010}
			& \best{0.4243}
			& \best{58.90} \\
			
			& S3Diff
			& 4.914
			& 0.3363
			& 0.2425
			& 35.57 \\
			
			& AdcSR
			& 6.540
			& 0.2652
			& 0.1871
			& 33.12 \\
			
			& Ours
			& 5.172
			& \second{0.5100}
			& \second{0.2579}
			& \second{46.95} \\
			
			\bottomrule
		\end{tabular}
	}
\end{table}

\subsection{Ablation Studies}
In this section, we conduct ablation studies to validate the effectiveness of our design choices for both the teacher and student models. It is worth noting that all metrics are computed on the DIOR test set.
\subsubsection{Effectiveness of Uncertainty Guided Adaptive Timestep}
\begin{figure}[htb]
	\centering
	\includegraphics[width=\linewidth]{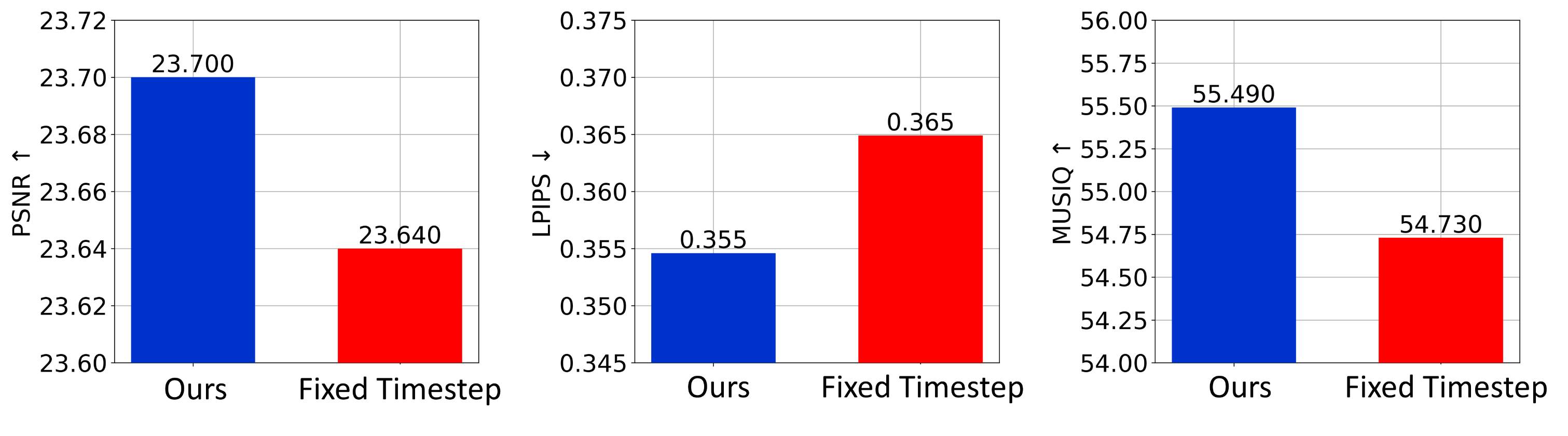}
	\caption{Ablation analysis of the uncertainty-guided strategy. For comparison, we retrain a simple single-step diffusion model with a fixed time step as the teacher model. The results show that the proposed teacher model achieves consistent improvements across all evaluation metrics.}
	\label{fig:abla_uncer}
\end{figure}

For the teacher model, we jointly leverage uncertainty information to estimate a adaptive timestep, which is then used to guide the single-step denoising process of the teacher UNet. To validate the effectiveness of this mechanism, we alternatively employ a single-step diffusion model with a fixed timestep as the teacher model and use it to supervise the student model via knowledge distillation. As shown in Fig.~\ref{fig:abla_uncer}, the proposed uncertainty-guided adaptive timestep mechanism consistently improves the performance of the student model. This is because, compared with a fixed timestep, the proposed teacher model can dynamically adjust the generative strength of the single-step diffusion model according to the spatial reconstruction difficulty of different inputs.

\begin{figure}[htb]
	\centering
	\includegraphics[width=\linewidth]{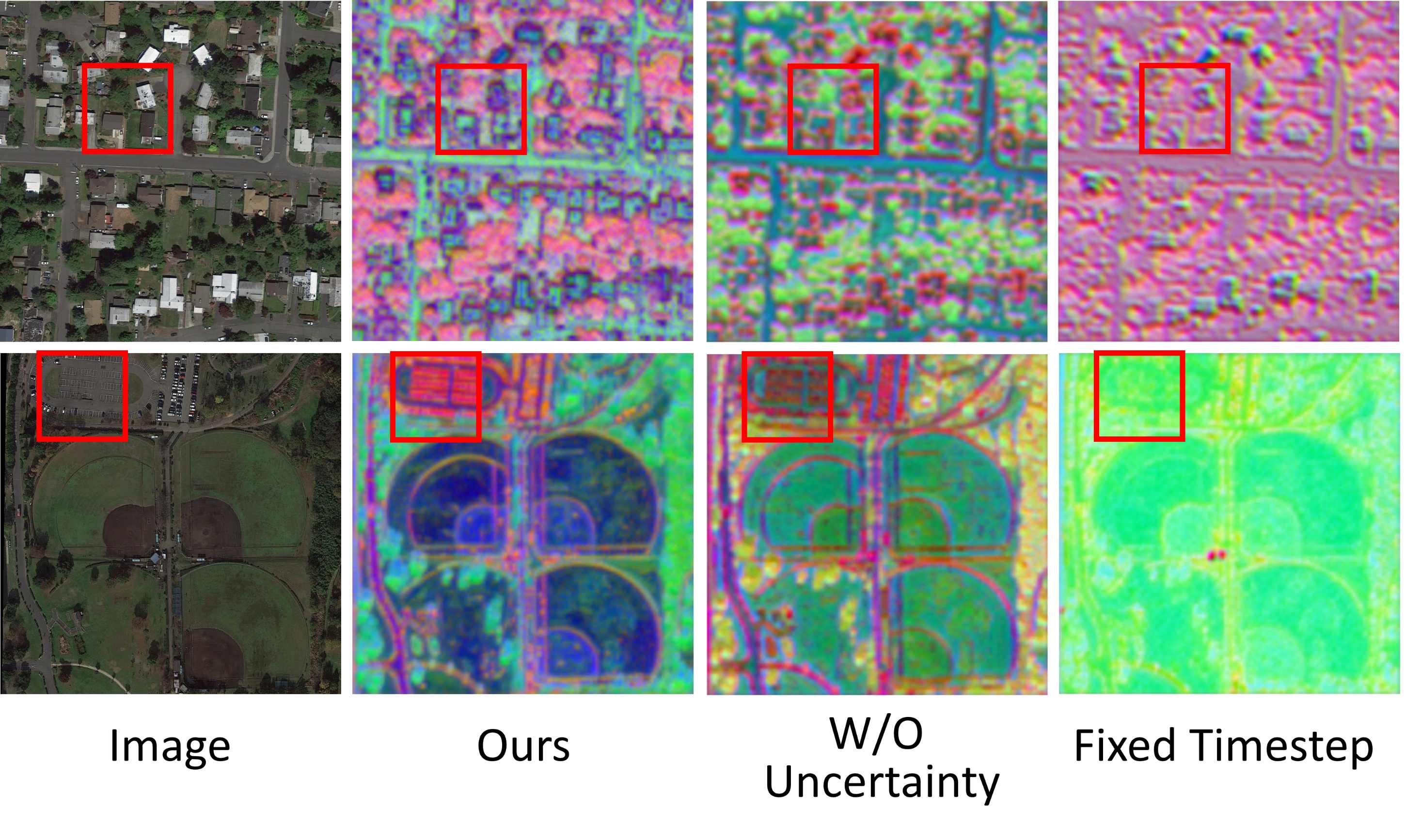}
	\caption{Visualization of intermediate features from the teacher U-Net. The proposed strategy that jointly combines uncertainty estimation with dynamic timestep assignment encourages the activation of higher-quality intermediate features, thereby providing more effective supervision for the student model.}
	\label{fig:feat_visual}
\end{figure}

In Fig.~\ref{fig:feat_visual}, we further visualize the intermediate features from the teacher UNet. First, we remove the uncertainty map and estimate the adaptive timestep using only the $z_{\mathrm{gt}}$ term in Eq.~\ref{eq:pred_alpha}. Second, we directly employ a fixed timestep for denoising. It can be observed that, compared with the two ablation variants, the proposed uncertainty guided teacher model is able to activate richer structural features, thereby providing higher-quality feature supervision for training the student model in the second stage. 

\begin{figure}[htb]
	\centering
	\includegraphics[width=\linewidth]{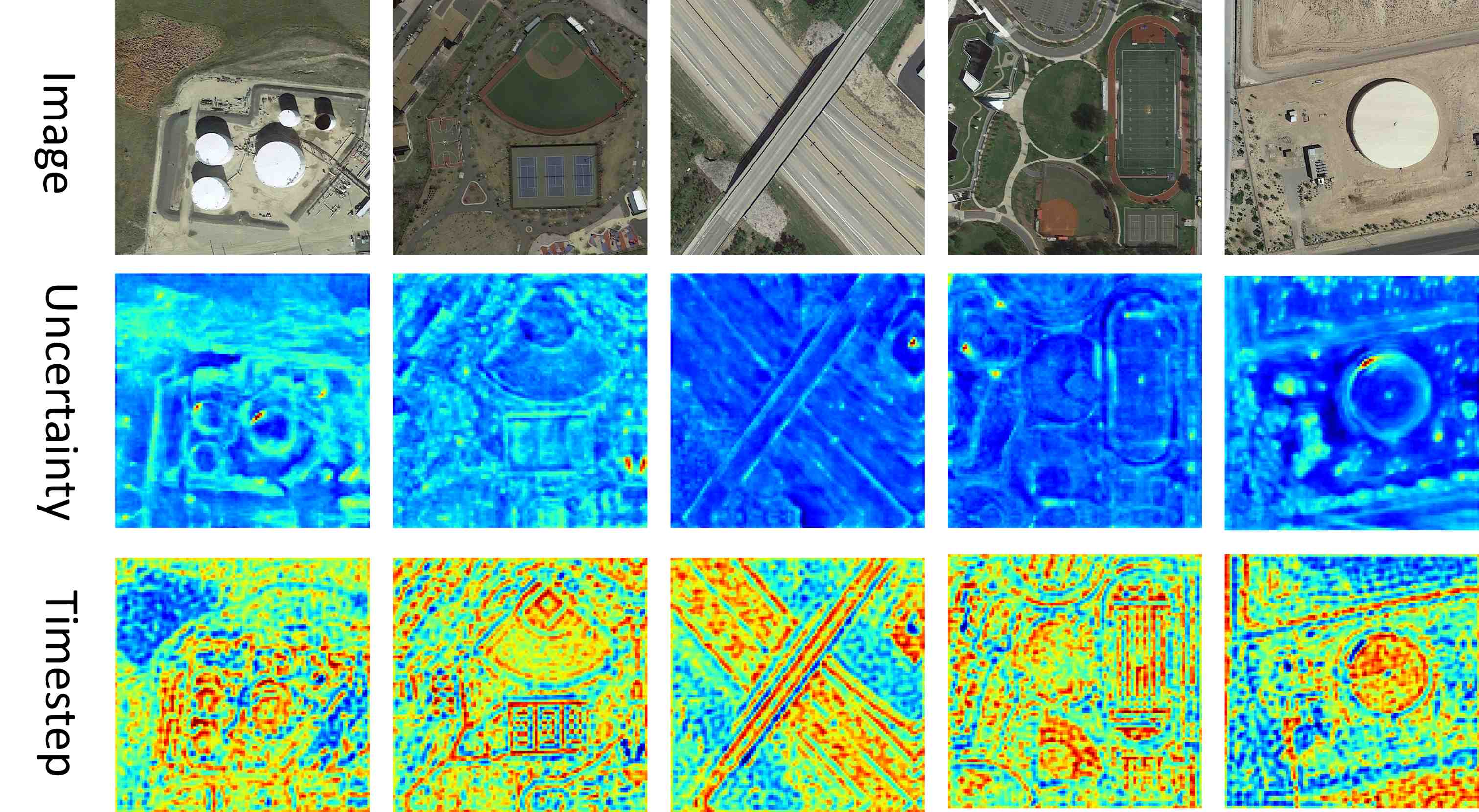}
	\caption{Visualization of uncertainty maps and the corresponding predicted timestep maps. The uncertainty map reflects reconstruction ambiguity, while the predicted timestep represents the learned adaptive guidance level of the teacher model. Although high-uncertainty regions generally tend to require stronger generative guidance, the timestep map is not a direct transformation of uncertainty, since it is jointly determined by uncertainty and latent structural context.}
	\label{fig:uncertainty}
\end{figure}

In Fig.~\ref{fig:uncertainty}, we present the uncertainty maps predicted by $\mathrm{UE}_{\phi}$. Regions with dense structures and rich textures usually exhibit higher uncertainty, indicating greater reconstruction ambiguity. We further visualize the predicted timestep maps. The results show that the timestep prediction is not always strictly monotonic with the uncertainty map. This is reasonable because the predictor is optimized by the final teacher restoration objective and also uses latent structural information during teacher construction. Therefore, in some low-uncertainty regions, larger timesteps may still be selected when stronger diffusion priors help improve structural consistency or perceptual restoration.

\subsubsection{Effectiveness of Enhanced Lightweight VAE}
To evaluate the effectiveness of our enhanced lightweight VAE, we compare it against the original large VAE and an extremely lightweight VAE variant (TAESD). The original VAE provides high reconstruction quality but is computationally expensive due to its large model size. TAESD is a lightweight alternative that significantly reduces parameters and inference time but may compromise reconstruction quality due to information bottlenecks in the latent space. Our enhanced VAE addresses this by fusing convolution layers between the VAE and UNet, alleviating the information bottleneck and preserving richer structural details while maintaining efficiency.

Table \ref{tab:abla_vae} presents the SR performance of the student model equipped with different variants of the VAE. Our method achieves the best MUSIQ score, indicating superior perceptual quality, while maintaining low inference time and moderate parameter count. Compared to the original VAE, it significantly reduces inference time by approximately 5 times and parameters by about 2.4 times, with only a slight drop in PSNR and LPIPS. Compared to TAESD, it improves PSNR by 0.38 and MUSIQ by 1.62, while keeping similar efficiency. This demonstrates that our enhanced VAE effectively balances reconstruction quality and computational efficiency.

\begin{table}[htb]
	\centering
	\caption{Ablation study on enhanced lightweight VAE. The best and second-best results are highlighted in \best{bold red} and \second{underlined blue}, respectively.}
	\label{tab:abla_vae}
	\resizebox{\columnwidth}{!}{
	\begin{tabular}{@{\extracolsep{\fill}}lccccc@{}}
		\toprule
		\multirow{2}{*}{Method} & \multicolumn{3}{c}{Image Quality Metrics} & \multicolumn{2}{c}{Efficiency Metrics} \\
		\cmidrule(lr){2-4} \cmidrule(lr){5-6}
		& PSNR $\uparrow$ & LPIPS $\downarrow$ & MUSIQ $\uparrow$ & Time (s) $\downarrow$ & Params (M) $\downarrow$ \\
		\midrule
		VAE & \best{23.95} & \best{0.3505} & \second{54.21} & 0.195 & 138.9 \\
		TAESD & 23.32 & 0.3621 & 53.87 & \best{0.037} & \best{57.68} \\
		Ours & \second{23.70} & \second{0.3546} & \best{55.49} & \best{0.037} & \second{58.05} \\
		\bottomrule
	\end{tabular}}
\end{table}

\subsubsection{Effectiveness of Module Removal Strategy}
To evaluate the effectiveness of our semantic-aware pruning (SAP) strategy, we compare it against the uniform channel pruning method adopted in AdcSR. As shown in Table \ref{tab:abla_modules}, our SAP strategy achieves superior performance by semantically identifying and removing redundant modules tailored to the SR task. The results demonstrate that simply removing channels uniformly (W/O SAP) leads to significantly higher parameter counts (168.3M) and inference time (0.052s), while substantially degrading perceptual quality (54.96 on MUSIQ). By contrast, our semantic-aware approach effectively reduces the model size while maintaining competitive perceptual quality, highlighting the importance of task-specific pruning strategies for remote sensing image SR.

\begin{table}[htb]
	\centering
	\caption{Ablation studies on module removal and module replacement strategies. The best and second-best results are highlighted in \best{bold red} and \second{underlined blue}, respectively.}
	\setlength{\tabcolsep}{3pt}
	\label{tab:abla_modules}
	\resizebox{\columnwidth}{!}{
	\begin{tabular}{@{\extracolsep{\fill}}lccccc@{}}
		\toprule
		\multirow{2}{*}{Method} & \multicolumn{3}{c}{Image Quality Metrics} & \multicolumn{2}{c}{Efficiency Metrics} \\
		\cmidrule(lr){2-4} \cmidrule(lr){5-6}
		& PSNR $\uparrow$ & LPIPS $\downarrow$ & MUSIQ $\uparrow$ & Time (s) $\downarrow$ & Params (M) $\downarrow$ \\
		\midrule
		W/O SAP & \best{23.89} & 0.3667 & 54.96 & 0.052 & 168.3 \\
		W/O DSC & 23.67 & 0.3621 & 55.28 & \second{0.040} & 93.18 \\
		W/O FSC & 23.44 & \second{0.3540} & 55.44 & 0.042 & 99.46 \\
		W/O QGAM & 23.34 & \best{0.3526} & \best{55.57} & 0.057 & \second{81.11} \\
		Ours & \second{23.70} & 0.3546 & \second{55.49} & \best{0.037} & \best{58.05} \\
		\bottomrule
	\end{tabular}}
\end{table}

% \begin{figure}[tb]
	%     \centering
	%     \includegraphics[width=0.85\linewidth]{fig/r4/ablation_radar_chart.png}
	%     \caption{Radar chart of different ablation variants. Our method achieves the highest inference efficiency while maintaining highly competitive reconstruction performance.}
	%     \label{fig:ablation_radar_chart}
	% \end{figure}

\subsubsection{Effectiveness of Module Replacement Strategy}
Beyond module removal, we further enhance efficiency through replacing standard convolutional and self-attention layers with our proposed lightweight alternatives. The purpose of these modules are not to surpass the original heavy modules in reconstruction accuracy, but to serve as lightweight substitutes that retain comparable SR performance with substantially lower model complexity. The ablation results in Table \ref{tab:abla_modules} progressively demonstrate the impact of each design choice. When removing the direction-separable convolution (DSC), PSNR drops to 23.67 and the model requires 93.18M parameters with 0.040s inference time. The frequency-separable convolution (FSC) contributes significantly to maintaining fine details, as removing it yields PSNR of 23.44 and slightly increases inference time to 0.042s. The query-driven global aggregation module (QGAM) provides crucial global context, and its absence results in the lowest PSNR (23.34) and highest inference time (0.057s) among the ablated variants. We observe that incorporating QGAM leads to slight decreases in LPIPS and MUSIQ. This behavior can be attributed to the global smoothing effect of the proposed global aggregation mechanism. Specifically, QGAM may reduce the diversity of high-frequency details, leading to less pronounced texture variations. Overall, these results demonstrate that the proposed module replacement strategy achieves a favorable trade-off between efficiency and reconstruction quality. This validates that RS-aware lightweight design can serve as an effective alternative to standard modules in diffusion-based SR models without incurring significant performance loss.
% As illustrated in Fig. \ref{fig:ablation_radar_chart}, we compare different ablation variants of our method using a radar chart. The results show that our approach can significantly improve inference efficiency with almost no degradation in reconstruction performance.

\begin{figure}[htb]
	\centering
	\includegraphics[width=\linewidth]{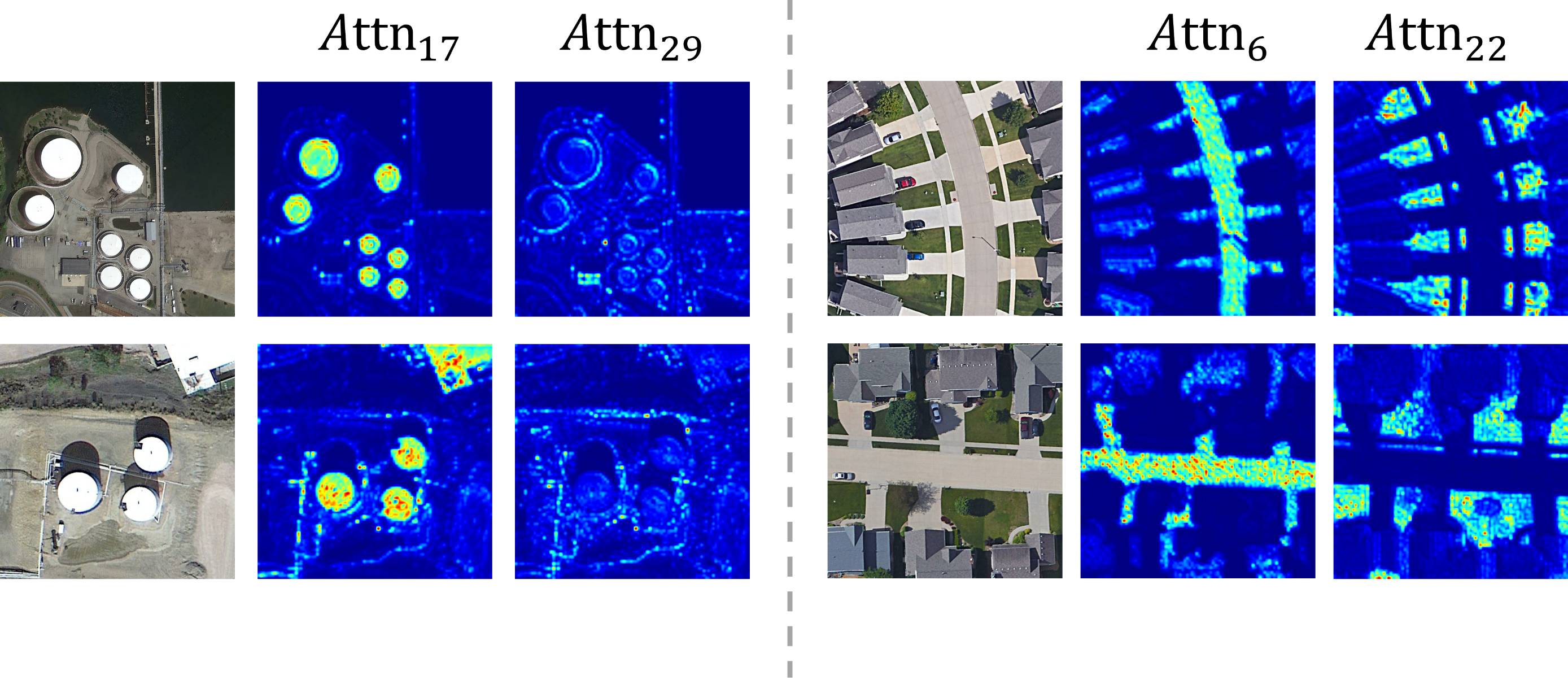}
	\caption{Visualization of the attention scores in the proposed QGAM module. $\text{Attn}_{N}$ denotes the attention map corresponding to the $N$-th learnable query token. It can be observed that, across different scenes, the learnable queries are able to aggregate semantically rich information and respond to specific land-cover types, such as storage tanks, background regions, roads, and grassland. Therefore, this design enables global token interaction with linear computational complexity.}
	\label{fig:atten_map}
\end{figure}

The proposed QGAM enables global token interaction through two cross-attention operations, effectively alleviating the high computational cost of global self-attention. Since remote sensing images are typically composed of a limited set of land-cover categories, each learnable query token in QGAM can act as a semantic proxy to extract specific types of information from the image. In Fig. \ref{fig:atten_map}, we visualize the attention maps corresponding to different learnable query tokens. It can be observed that, across various scenes, these queries are able to aggregate rich semantic information and respond to specific land-cover types, such as storage tanks, background regions, roads, and grassland. Therefore, QGAM serves as an efficient alternative to computationally expensive self-attention modules while maintaining lightweight inference efficiency.

\begin{table}
	\centering
	\caption{Ablation study on knowledge distillation losses. The best and second-best results are highlighted in \best{bold red} and \second{underlined blue}, respectively.}
	\label{tab:abla_dis}
	\begin{tabular*}{\columnwidth}{@{\extracolsep{\fill}}lccc@{}}
		\toprule
		Method & PSNR $\uparrow$ & LPIPS $\downarrow$ & MUSIQ $\uparrow$ \\
		\midrule
		No-Dis & 22.69 & 0.4370 & \best{57.66} \\
		No-MMD & \second{23.47} & \second{0.3574} & 53.99 \\
		Ours & \best{23.70} & \best{0.3546} & \second{55.49} \\
		\bottomrule
	\end{tabular*}
\end{table}

\begin{figure}[htb]
	\centering
	\includegraphics[width=\linewidth]{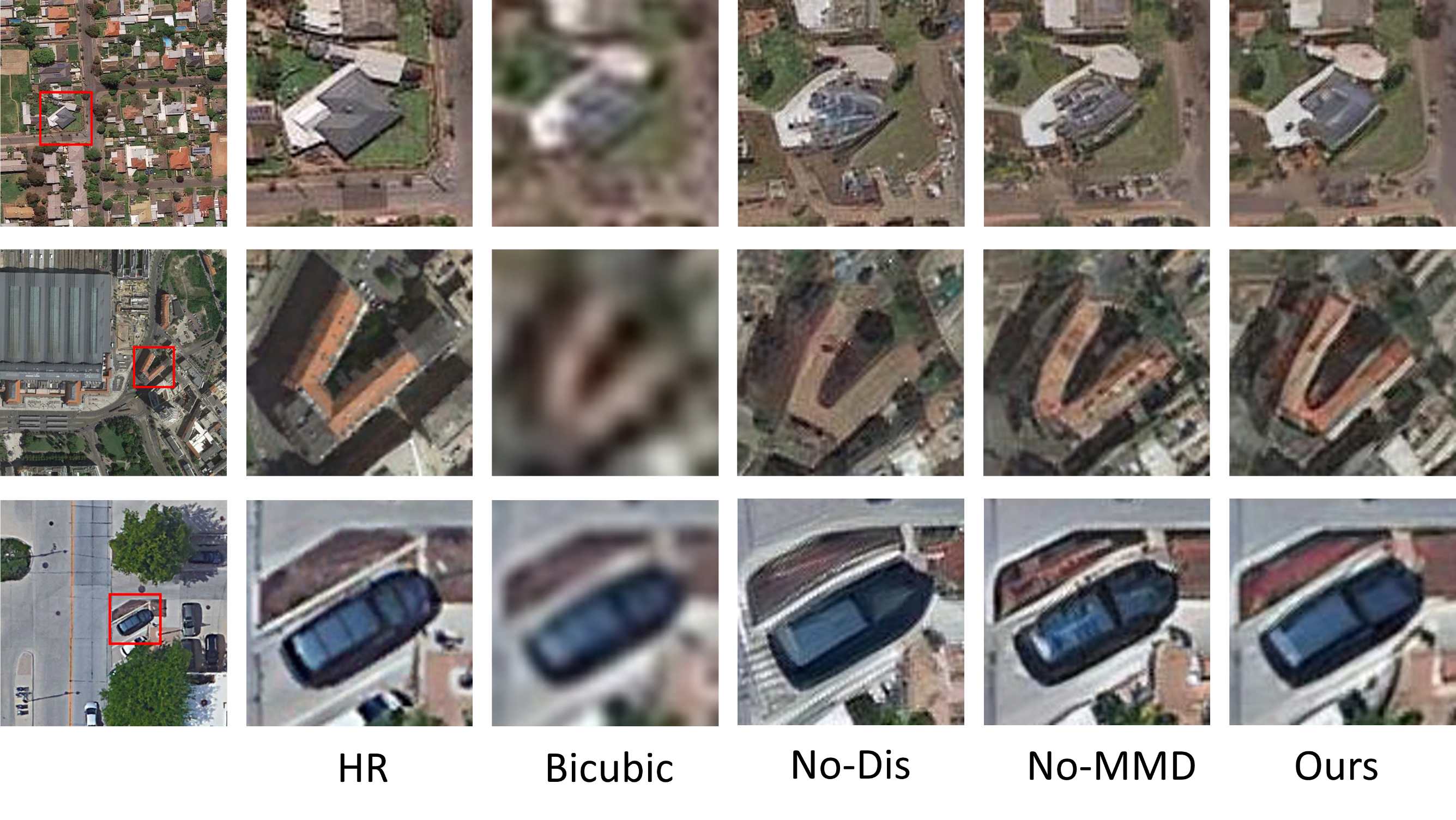}
	\caption{Visual comparison for the ablation study on knowledge distillation. The inclusion of the distillation loss and the MMD loss significantly improves the semantic fidelity of the reconstructed results.}
	\label{fig:visual_abla}
\end{figure}

\subsubsection{Effectiveness of Knowledge Distillation}
To verify the contribution of the distillation objective, we compare three variants:
\textbf{No-Dis} (without any distillation loss), \textbf{No-MMD} (using only MSE distillation), and \textbf{Ours} (using both MSE and MMD losses). As shown in Table \ref{tab:abla_dis}, introducing distillation significantly improves reconstruction fidelity. Compared with No-Dis, the No-MMD variant improves PSNR from 22.69 to 23.47 and reduces LPIPS from 0.4370 to 0.3574, demonstrating that feature-level supervision from the teacher model is essential for preserving faithful structures and textures. It is worth noting that the No-Dis variant achieves a relatively high MUSIQ score. This is because, without any feature-level constraints, the model tends to generate excessive textures. However, these textures are often unrealistic and inconsistent with the ground truth, which leads to a significant degradation in full-reference metrics such as PSNR and LPIPS.

Furthermore, adding the proposed MMD loss on top of MSE leads to consistent gains across all metrics. Specifically, our full method further improves PSNR to 23.70, reduces LPIPS to 0.3546, and increases MUSIQ to 55.49 compared with No-MMD. These results indicate that MSE-based point-wise alignment alone is insufficient to fully transfer teacher knowledge, while the additional MMD constraint effectively reduces the distribution gap between teacher and student features. Consequently, the student model can achieve a better balance between perceptual quality and reconstruction fidelity without introducing additional inference overhead. In Fig. \ref{fig:visual_abla}, we present a visual comparison of reconstruction results from different ablation variants. As observed in the first two rows, the introduction of the distillation loss and the MMD loss significantly improves the semantic consistency of the reconstruction results, as evidenced by clearer building boundaries. From the last row, it can be seen that our method effectively reduces artifacts in the SR images. These results further demonstrate the effectiveness of the proposed knowledge distillation strategy.
\section{Conclusion}
In this paper, we propose SlimDiffSR, a lightweight and efficient diffusion framework for real-world remote sensing image super-resolution. The method follows a two-stage paradigm, where an uncertainty-guided single-step teacher model is first constructed, followed by a compact student model obtained via structured pruning and knowledge distillation. Specifically, we introduce dynamic timestep assignment through uncertainty estimation and timestep inversion, develop an enhanced lightweight VAE and a semantic-aware UNet pruning strategy, and incorporate direction- and frequency-separable operators together with a query-driven global aggregation module to improve efficiency. Moreover, we combine MSE and MMD losses to enhance distribution-level feature alignment during distillation, enabling the student model to better approximate the teacher.

Extensive experiments on multiple remote sensing benchmarks demonstrate that SlimDiffSR achieves a favorable balance between reconstruction quality and efficiency, delivering strong perceptual performance with significantly reduced inference cost compared to existing diffusion-based methods. Ablation studies further validate the effectiveness of each component. While the method remains competitive under more challenging settings such as $8\times$ SR and real-world SR, improving robustness under severe degradations and cross-sensor domain gaps remains an important direction. Future work will explore stronger degradation-aware priors, arbitrary-scale SR, and more general lightweight diffusion frameworks for large-scale Earth observation applications.
% \textcolor{red}{red}
{
    \small
    \bibliographystyle{ieeenat_fullname}
    \bibliography{main}
}

\end{document}